\documentclass[fleqn,11pt]{wlscirep}

\usepackage[T1]{fontenc}
\usepackage{setspace}
\usepackage{multirow}
\usepackage{subcaption}
\usepackage{array}

\newcolumntype{R}[1]{>{\raggedleft\arraybackslash}p{#1}}
\newcolumntype{L}[1]{>{\raggedright\arraybackslash}p{#1}}
\newcolumntype{P}[1]{>{\centering\arraybackslash}p{#1}}

\title{Video-based detection of cessation of breathing in pre-term infants using machine learning}
\author[1]{Dineo Serame}
\author[1]{Lionel Tarassenko}
\author[1]{Mauricio Villarroel}
\affil[1]{Institute of Biomedical Engineering, Department of Engineering Science,  University of Oxford, UK}
\affil[*]{mauricio.villarroel@eng.ox.ac.uk}

\begin{abstract}
Pre-term infants are susceptible to potentially harmful apnoea-related cessations of breathing due to immature respiratory control mechanisms. However, reliable respiratory monitoring in this cohort remains challenging in the neonatal intensive care unit (NICU), where motion artefacts, sensor displacement, and skin fragility can compromise traditional contact-based measurements. Non-contact video monitoring offers a potential complementary modality that does not depend on adhesive sensors and may provide additional respiratory information.

\paragraph{}
We evaluate whether camera-based signals can detect apnoea-related cessation of breathing (COBE) and whether they provide complementary information to routinely acquired physiological signals. Using video and clinical recordings from 30 pre-term infants, respiratory motion was extracted from dynamically tracked torso regions to generate camera-derived time-series signals. Camera-only models were trained using residual network (ResNet) architectures. Hybrid models subsequently combined video-derived signals with impedance pneumography (IP), ECG-derived respiration (EDR) and the PPG-derived respiratory envelope.  

\paragraph{}
Camera-only models achieved a balanced accuracy of 76.9\%, demonstrating the feasibility of non-contact detection of COBE events. Fusion of video-derived features with IP improved balanced accuracy to 90.6\%, outperforming either modality alone and indicating that video-derived features contribute additional respiratory information beyond standard physiological signals.

\paragraph{}
These findings show that video-derived signals contain clinically relevant respiratory features and can enhance COBE detection performance when combined with conventional physiological signals. This supports the use of non-contact video as a complementary modality for automated COBE detection and highlights its potential to improve the robustness of neonatal respiratory monitoring.
\end{abstract}

\begin{document}
\setlength{\belowdisplayskip}{10pt} \setlength{\belowdisplayshortskip}{5pt}
\setlength{\abovedisplayskip}{1pt} \setlength{\abovedisplayshortskip}{3pt}

\raggedbottom
\maketitle

\maketitle

{\normalfont
\noindent\textbf{Keywords:}
Apnoea of prematurity; neonatal intensive care;
video-based respiratory monitoring;
non-contact monitoring;
machine learning;
multimodal data fusion.
}

\vspace{1em}

\thispagestyle{empty}
\newpage

\section{Introduction}
\label{intro}
Pre-term birth, defined as delivery before 37 weeks of gestation, remains a leading cause of neonatal morbidity and mortality worldwide~\cite{world2023born,koteswari2022preterm,platt2014outcomes}. Respiratory complications such as apnoea, driven by immature respiratory control and lung development, are major contributors to mortality in this population. Apnoea of prematurity (AOP) affects up to 85\% of infants born at or before 34 weeks of gestation~\cite{thompson2024apnea,o2025caffeine}. Episodes of cessation of breathing (COBE) are a defining feature of AOP and may result in hypoxaemia (reduced oxygen levels in the blood), bradycardia, neurological injury, and long-term developmental complications, including cognitive impairment and motor dysfunction~\cite{mohr2015very,pergolizzi2022epidemiology}. Accurate and reliable detection of COBE episodes is therefore critical for timely clinical intervention and improved outcomes.

\paragraph{}
Although polysomnography remains the gold standard for diagnosing apnoea~\cite{zou2023new,nwaneri2024review}, it is impractical for continuous monitoring in the neonatal intensive care unit (NICU) due to its complexity, requirement for specialised equipment, and need for trained personnel. Routine respiratory monitoring and apnoea detection in the NICU rely primarily on contact-based techniques, including impedance pneumography (IP) and pulse oximetry~\cite{pullano2017medical,bertoni2019towards}. Impedance pneumography uses chest electrodes to measure changes in transthoracic electrical impedance that occur due to cyclic changes in lung air volume during breathing~\cite{bawua2021review,massaroni2019contact}. These impedance variations provide an indirect measure of respiratory effort and are commonly used to estimate respiratory rate (RR) and identify respiratory pauses. Pulse oximetry is an optical technique used to estimate peripheral oxygen saturation ($SpO_{2}$) and identify desaturation events associated with COBE episodes~\cite{poets2003pulse}.

\paragraph{}
Despite their widespread use, contact-based monitoring systems are subject to several limitations. Motion artefacts, electrode displacement, and cardiac-related fluctuations in impedance signals can generate false alarms or obscure respiratory pauses~\cite{bawua2021review,poets2003pulse,ostojic2020reducing}. The attachment and removal of adhesive sensors may also contribute to skin irritation, discomfort, and epidermal injury in pre-term infants with immature and highly sensitive skin~\cite{baharestani2007overview,sharma2025chronological}. Contact sensors can also interfere with routine caregiving practices such as kangaroo care, which promotes physiological stability and bonding~\cite{lloyd2015overcoming,bonner2017there}. These constraints are particularly pronounced in low-resource settings, where limited monitoring infrastructure and a shortage of clinical staff may delay timely recognition of respiratory events~\cite{moxon2015inpatient,kamala2022availability}. Collectively, these challenges highlight the need for monitoring approaches that enhance robustness while reducing physical burden.

\paragraph{}
Non-contact physiological monitoring has emerged as a promising alternative approach. Technologies such as thermal imaging, Doppler radar, WiFi-based sensing, and video camera-based monitoring can capture respiratory-related signals without direct skin contact~\cite{cattani2014wire,yang2016sleep,muhammad2019p016}. Thermal cameras detect respiration-induced temperature fluctuations near the airway~\cite{fei2009thermal}. Doppler radar systems estimate respiration by tracking chest or abdominal wall movements during breathing~\cite{ishrak2023doppler}. WiFi-based methods infer respiration from breathing-induced variations in wireless signal propagation characteristics~\cite{michaelis2025vitalcsi}. While these modalities reduce sensor-related discomfort, they may be costly, susceptible to environmental interference, and remain insufficiently validated in neonatal populations~\cite{vitazkova2024advances,maurya2021non}. 

\paragraph{}
Visible-spectrum (RGB) video cameras offer a comparatively accessible and scalable approach. Subtle thoracic or abdominal motion can be extracted through image processing and machine learning techniques~\cite{geertsema2020automated,villarroel2019non,villarroel2020non}. In adult cohorts, video-based methods have achieved high sensitivity for respiratory event detection under controlled conditions~\cite{akbarian2020distinguishing,pinilla2025diagnostic, jorge2022non,villarroel2020non}. However, translating video-based respiratory monitoring to pre-term infants introduces distinct technical challenges. Neonatal respiratory patterns are more irregular than in adults, with breath-to-breath variability, making RR estimation more challenging~\cite{coleman2022assessment,adjei2021new}. Cardiac activity can also introduce artefacts into respiratory signals by superimposing cardiac-induced fluctuations, reducing the accuracy of respiratory signal extraction~\cite{lee2011new,lorato2021video}. Unlike adult sleep studies, AOP may occur across sleep-wake states, including wakefulness and rapid eye movement (REM) sleep, requiring respiratory monitoring across a broader range of physiological states~\cite{zhao2011apnea,haskova2014apnea,joosten2017sleep,levy2017impact}. These physiological challenges are compounded by the NICU environment, where infants are frequently repositioned or partially occluded by bedding or caregivers, and incubator structures can introduce reflections and visual interference~\cite{selvaraju2022continuous}. Variable lighting and spontaneous limb movements further complicate feature extraction, requiring methods that are robust to real-world NICU conditions. Most prior neonatal studies rely on small sample sizes, observer-dependent annotations, or controlled laboratory settings, limiting generalisability to routine clinical environments~\cite{cattani2014wire,lorato2021video}.

\paragraph{}
Contact-based sensors quantify physiological parameters such as RR and $SpO_{2}$ but do not capture visual context such as posture changes, caregiver interactions, or transient motion patterns. Vision-based monitoring may provide complementary information by capturing surface motion cues that help distinguish respiratory pauses from artefactual fluctuations in physiological signals. Integrating contact and non-contact modalities may therefore improve the robustness of COBE event detection by combining physiological measurements with contextual visual information. However, the extent to which video-derived information provides complementary value beyond conventional physiological monitoring for COBE detection remains unclear.
 
\paragraph{}
In this study, we propose video camera-based and hybrid frameworks for COBE detection in pre-term infants. Using synchronised video and physiological recordings acquired under routine NICU conditions, we evaluate whether video-derived signals contain clinically relevant information for COBE detection and whether their integration with conventional physiological measurements improves detection performance.

\section{Methods}
\label{methods}
\subsection{Clinical cohort and data acquisition}
We used data from a clinical study of pre-term infants admitted to the NICU at the John Radcliffe Hospital, Oxford, UK. The study was conducted in a collaboration between Oxford University Hospitals NHS Foundation Trust and the Oxford Biomedical Research Centre (BRC), with approval from the South Central - Oxford Research Ethics Committee (13/SC/0597). Written informed consent was obtained from parents prior to participation. 

\begin{figure}[!h]
\centering
\includegraphics[width = 18.0 cm]{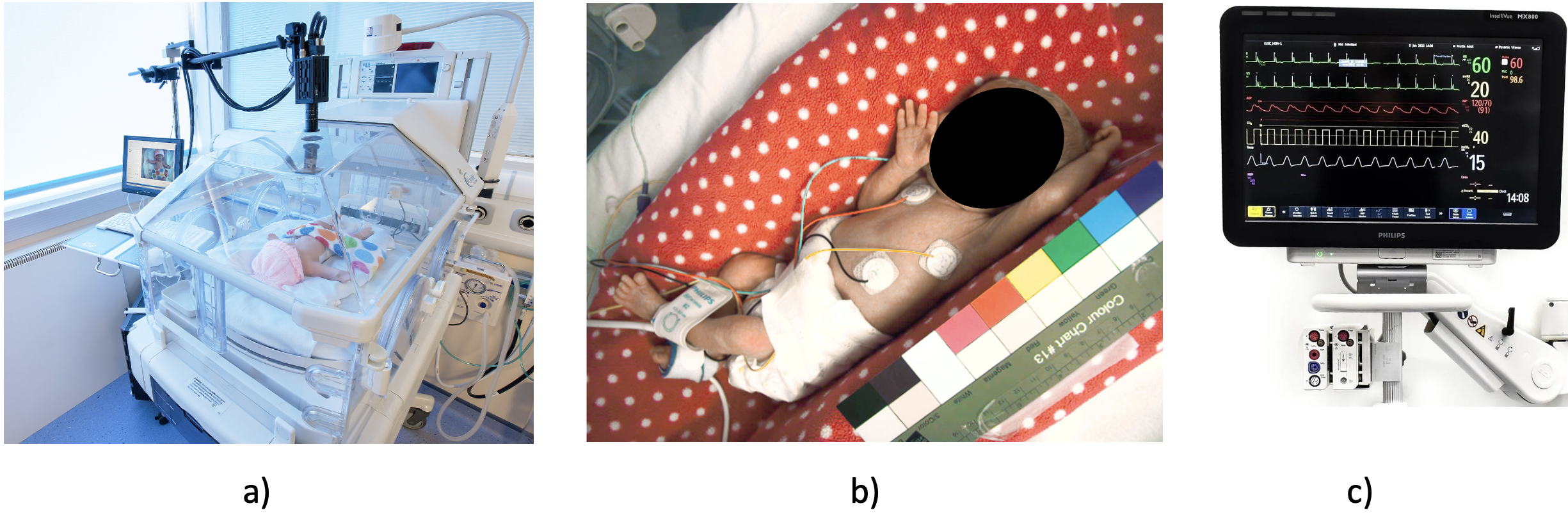}
    \caption{Components of the data acquisition setup used for synchronised video and physiological monitoring in the NICU. a) Video camera positioned through a modified opening in the incubator canopy; b) Representative video frame acquired during monitoring; c) Representative Philips patient monitor with $SpO_2$ measurement modules.}    
         \label{babycam}
    \end{figure}

\paragraph{}
Infants born at less than 37 weeks gestation and requiring high-dependency care were eligible for inclusion. The cohort comprised 30 pre-term infants (mean gestational age 31.1 $\pm$ 1.8 weeks) monitored for up to 7 consecutive days during daytime under routine clinical conditions. There were a total of across 90 recording sessions, yielding approximately 426 hours of synchronised video and physiological recordings. A 3-CCD JAI AT-200CL digital camera (JAI A/S, Denmark) was positioned through a modified opening in the canopy of a Giraffe OmniBed Carestation incubator (General Electric, Connecticut, USA) (figure~\ref{babycam}a). The camera captured continuous 24-bit RGB video at 20 frames per second with a resolution of 1628 $\times$ 1236 pixels. A representative video frame is shown in figure~\ref{babycam}b. Standard physiological signals, including heart rate (HR), respiratory rate (RR), and $SpO_2$, were recorded using a Philips IntelliVue MX800 patient monitor (Philips, Amsterdam, Netherlands) shown in figure~\ref{babycam}c. The monitor was equipped with a Masimo SET $SpO_2$ Vuelink IntelliVue measurement module (Masimo, California, USA) for oxygen saturation monitoring . Video acquisition was temporarily paused during selected clinical procedures, including phototherapy, cannula insertion, or kangaroo care. The full cohort design and data acquisition protocol have been described previously in Villarroel \textit{et al.}~\cite{villarroel2019non}.

\subsection{Recomputing reference RR}
\begin{center}
\begin{figure}[h!]
\centering
\includegraphics[width = 17.0 cm]{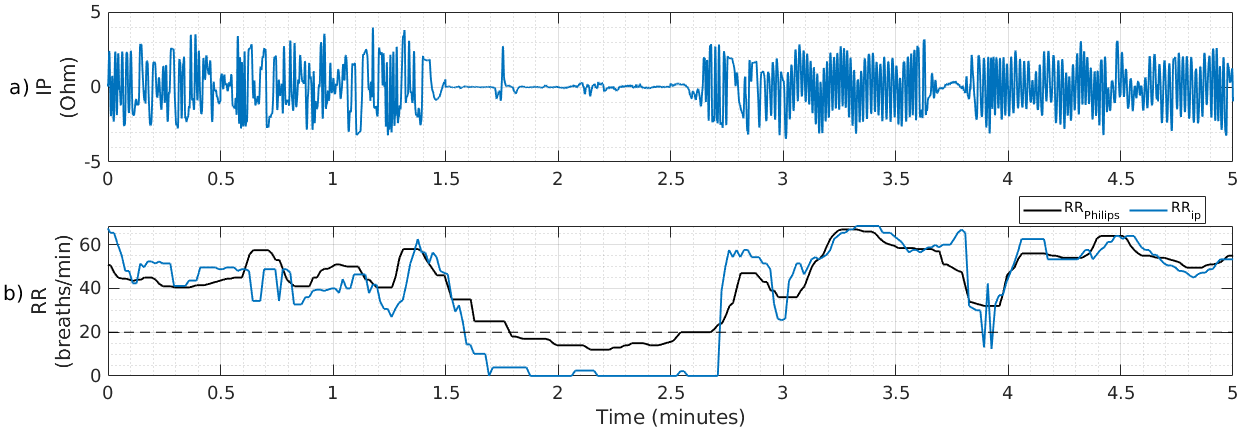}
    \caption{RR estimation using the breath counting method on a 5-minute segment of the IP signal containing a COBE episode (approximately 1.6--2.7 mins):  a) Filtered IP waveform;  b) comparison of the recomputed respiratory rate ($RR_{ip}$) and the monitor-provided respiratory rate ($RR_{philips}$) during a period containing COBE. $RR_{philips}$ exhibits reduced short-term variability, consistent with temporal averaging.}
         \label{estimated_rrip}
    \end{figure}
    \end{center}

Previous analysis on our dataset demonstrated substantial discrepancies between monitor-provided respiratory rate ($RR_{philips}$) and manual breath counting, indicating that the monitor values were insufficiently reliable for use as a reference signal~\cite{8681397}. In addition, the Philips IntelliVue MX800 derives respiratory rate using proprietary signal-processing algorithms and does not provide a corresponding measure of signal quality~\cite{chaichulee2018non}. While the processing pipeline and averaging window are not publicly documented, monitor-provided RR estimates exhibited reduced short-term variability (figure~\ref{estimated_rrip}b). Such behaviour is consistent with temporal averaging approaches commonly used in bedside monitoring systems to provide stable respiratory measurements. These approaches may reduce sensitivity to transient respiratory fluctuations and short respiratory pauses such as those during COBE episodes~\cite{adjei2021new,honda2023effect}. Therefore, respiratory rate was recomputed directly from the raw IP waveform ($RR_{ip}$).

\paragraph{}
To remove baseline drift and high-frequency noise, the IP signal was bandpass filtered using an 8th-order high-pass Butterworth filter (0.033 Hz) and a 6th-order low-pass Butterworth filter (2.83 Hz), preserving respiratory frequencies corresponding to approximately 2--170 breaths/min. Respiratory cycles were then identified using the Mean Average Curve (MAC) peak detector~\cite{lu2006semi}, and respiratory rate was estimated using breath counting within a 10-second sliding window with a 1-second step size. Signal quality assessment was applied to individual respiratory cycles based on physiological plausibility, spectral characteristics, and waveform purity. Respiratory cycles that failed these quality criteria were excluded from RR estimation. Figure~\ref{estimated_rrip} shows RR estimation over a 5-minute segment containing a period of cessation of breathing. Full details of the RR recomputation procedure are provided in Appendix~\ref{appendix:rr_ip}.

\subsection{Reference labelling of COBE events}
COBE labels were derived from physiological recordings according to established neonatal clinical criteria for apnoea~\cite{zhao2011apnea,pergolizzi2022limited}. A COBE event was defined as either (i) a respiratory pause ($RR_{ip}$ < 20 breaths/min) lasting at least 20 seconds, or (ii) a shorter respiratory pause of at least 10 seconds accompanied by bradycardia (HR < 100 beats/min) or oxygen desaturation ($SpO_2$ < 80\% lasting at least 10 seconds). 

\paragraph{}
Potential candidate COBE events were identified through automated screening of periods where oxygen saturation ($SpO_{2}$) remained below 80\% for at least 10 seconds. Consecutive desaturation episodes occurring within 20 seconds of each other were merged to avoid splitting prolonged physiological events into multiple candidate segments because of brief transient recoveries in oxygen saturation. This approach is consistent with observations that apnoea-related events may occur in temporal clusters during periods of physiological instability~\cite{joshi2016pattern}. This procedure identified 620 candidate desaturation-associated segments. To ensure representation of COBE events occurring both with and without oxygen desaturation, an additional 620 non-desaturation segments were randomly sampled from the same recording sessions for manual review. 

\paragraph{}
Each candidate segment was independently reviewed by three annotators (one clinician and two biomedical engineers) using $RR_{ip}$, $SpO_2$, ECG, PPG, and IP waveforms. The reviewers examined 5-minute contextual windows centred on each segment to determine whether a COBE episode had occurred. The decision workflow used to standardise annotation is shown in Appendix~\ref{appendix:supp_dataset}.

\paragraph{}
Confirmed COBE events were extracted as 80-second segments comprising 60 seconds preceding and 20 seconds following the onset of the annotated event, providing temporal context before and after the event. Normal breathing events were extracted as 80-second segments centred on the midpoint of the corresponding 5-minute windows labelled as normal breathing. Segments containing missing physiological data resulting from sensor disconnection, or for which the three reviewers did not agree, were excluded from further analysis. The final dataset comprised 346 COBE segments and 608 normal breathing segments derived from 24 infants. Inter-rater reliability was substantial for desaturation-associated events (Fleiss’ $\kappa = 0.80$) and moderate for non-desaturation events ($\kappa = 0.57$), reflecting the greater ambiguity in identifying respiratory pauses in the absence of associated desaturation. 

\subsection{Dynamic ROI selection}
 The annotated 80-second segments were processed using an automated region-of-interest (ROI) detection pipeline based on MediaPipe Pose, which uses the BlazePose pose-estimation model~\cite{bazarevsky2020blazepose}. Although BlazePose is primarily trained on adult human data, it was applied to estimate anatomical landmarks in neonatal recordings for dynamic ROI localisation. 

\paragraph{}
BlazePose identifies 33 anatomical landmarks corresponding to major body joints. For ROI definition, only the left and right shoulders and hips were utilised. These landmarks were used to define two regions of interest, illustrated in figure~\ref{landmarks}: (i) a torso ROI capturing global body motion, and (ii) a respiratory ROI (RR ROI) targeting localised breathing-related motion.  

\subsubsection{Torso ROI}
The torso centre was computed as the midpoint between the shoulder midpoint and the hip midpoint. A quadrilateral defined by the shoulder and hip landmarks was used to estimate torso orientation as shown in figure~\ref{landmark_success}. A minimum-area bounding rectangle aligned with the torso axis was fitted and expanded by 10\% to accommodate variations in posture. 

\paragraph{}
To preserve gradual motion while reducing abrupt frame-to-frame shifts in the estimated landmark positions, the bounding box centre was smoothed using a recursive exponential moving average (EMA) filter defined as

\begin{equation}
\hat{\mathbf{c}}_{t}
= \alpha_{t}\mathbf{c}_{t}
+ (1-\alpha_{t})\hat{\mathbf{c}}_{t-1}.
\end{equation}

where $\mathbf{c}_{t}$ is the current bounding box centre, $\hat{\mathbf{c}}_{t}$ is the smoothed centre, and $\alpha_{t}$ is an adaptive smoothing coefficient. The smoothing coefficient ($\alpha_t$) was adapted according to the frame-to-frame displacement of the bounding box centre. The displacement thresholds and corresponding smoothing coefficients were selected empirically based on visual inspection of ROI stability across representative neonatal recordings. The frame-to-frame displacement was defined as

\begin{equation}
d_t = \left\| \mathbf{c}_t - \hat{\mathbf{c}}_{t-1} \right\|_2,
\end{equation}

where $d_t$ is the Euclidean distance (in pixels) between the current bounding box centre ($\mathbf{c}_t$) and the smoothed centre from the previous frame ($\hat{\mathbf{c}}_{t-1}$). The smoothing coefficient was then selected according to

\begin{equation}
\alpha_t=
\begin{cases}
0.2, & d_t \leq 2.55,\\
0.6, & 2.55 < d_t \leq 7,\\
1.0, & d_t > 7.
\end{cases}
\end{equation}

Lower smoothing coefficients placed greater emphasis on the previous bounding box centre to suppress landmark localisation jitter, whereas higher coefficients increased the influence of the current landmark estimate, allowing rapid ROI repositioning during substantial body motion. The resulting rectangle constituted the torso ROI and captured large-scale torso motion.

\subsubsection{RR ROI}
\begin{figure}[h!]
 \centering
 \begin{subfigure}[b]{0.45\textwidth}
 \centering
 \includegraphics[width=\textwidth]{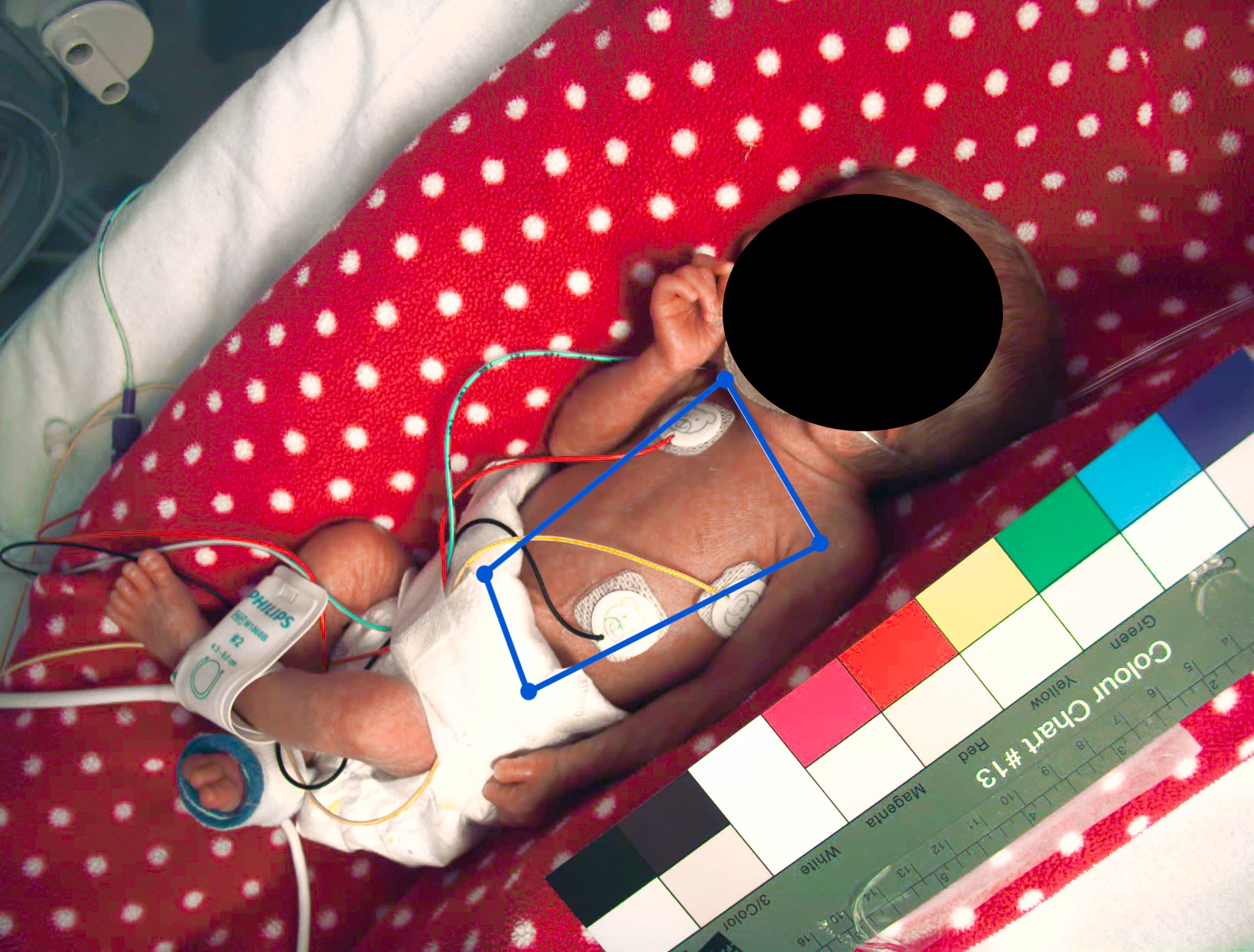}
 \caption{}
 \label{landmark_success}
 \end{subfigure}
  \hspace{0.01\textwidth} 
 \begin{subfigure}[b]{0.45\textwidth}
 \centering
 \includegraphics[width=\textwidth]{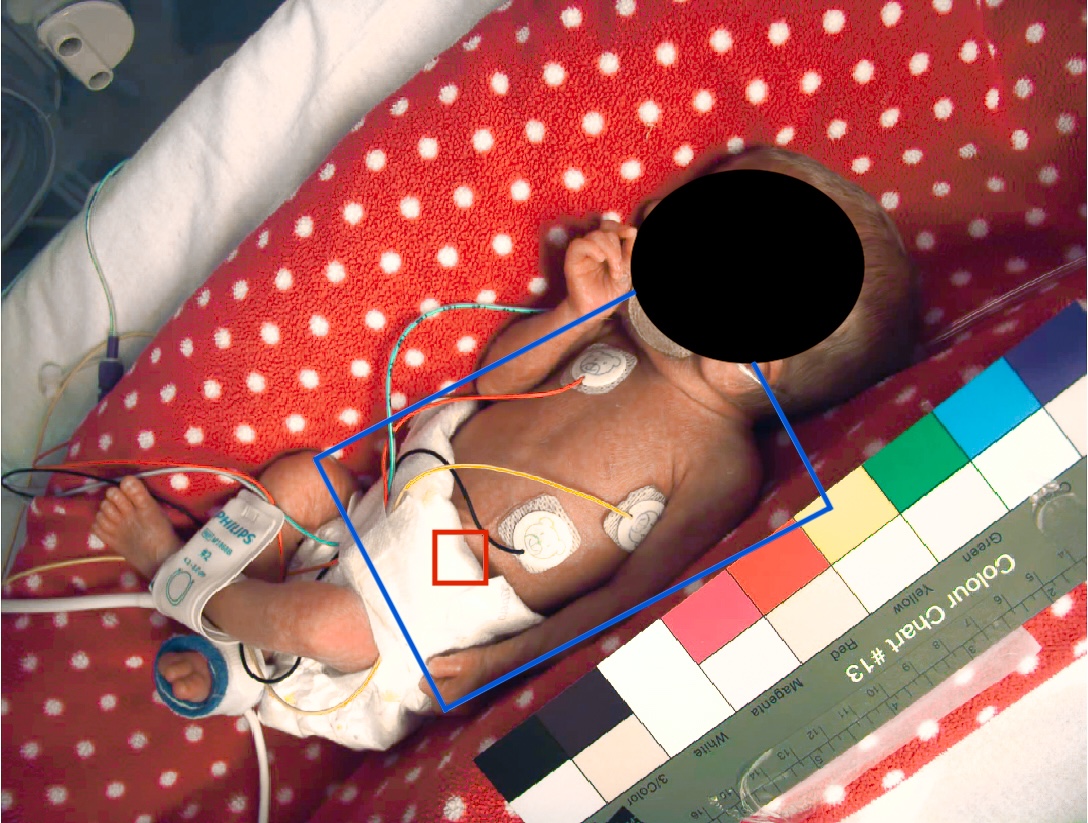}
 \caption{}
 \label{ROIs}
 \end{subfigure}
\caption{Dynamic ROI localisation using BlazePose landmarks. (a) Left and right shoulder and hip landmarks (blue circular markers) connected to form a torso quadrilateral used to estimate torso orientation; (b) The resulting torso ROI (blue). A $75 \times 75$ pixel respiratory ROI (red) was positioned over the lower abdominal region to capture respiratory-related motion.}
 \label{landmarks}
\end{figure}

A $75 \times 75$ pixel RR ROI was positioned over the lower abdominal region where breathing-induced motion is typically most pronounced. The RR ROI position was updated dynamically and smoothed using the same recursive filter applied to the torso ROI to reduce frame-to-frame jitter. The ROI size was selected based on previous work showing that small ROIs are more likely to contain breathing-related motion~\cite{chaichulee2018non,montoya2017non}. Because respiratory movements are spatially localised, a small ROI focused on the abdominal region reduces interference from non-respiratory body movements and background motion. Figure~\ref{ROIs} illustrates the resulting torso and RR ROIs.

\subsection{Camera-derived signal processing}

\subsubsection{Video camera-derived signal extraction}
Using the dynamically tracked torso and respiratory ROIs, two camera-derived time-series signals were extracted from each 80-second video segment. First, a frame-difference (FD) signal was computed from the torso ROI to quantify global infant motion. This signal was computed as the mean absolute pixel-wise intensity difference between consecutive frames within the torso ROI. The resulting FD signal captures both respiratory motion and non-respiratory body movement. Second, a pixel-intensity-based respiratory signal ($PPGi_{rr}$) was extracted from the respiratory ROI. For each frame, the mean pixel intensity within the $75 \times 75$ ROI was computed, generating a time-series that reflects respiratory-induced abdominal motion. Figures~\ref{cam_sig_extracted} presents representative examples of the extracted signals alongside physiological waveforms.

\subsubsection{Segment quality assessment}
All extracted neonatal video segments were visually reviewed to identify recordings unsuitable for respiratory analysis. Segments were excluded if the infant's torso was obstructed from the camera view during clinical interventions or kangaroo care, or when the infant was moved out of frame. 

\paragraph{}
Of the annotated 346 COBE segments, 100 were excluded, leaving 246 COBE segments suitable for camera-based analysis.  Similarly, 165 of the 608 normal breathing segments were excluded, resulting in 443 retained normal breathing segments. The final dataset therefore comprised 246 COBE episodes and 443 normal breathing episodes from 23 infants, for a total of 689 80-second segments. 

\subsection{Construction of the machine learning dataset}
\begin{figure}[!h]
\centering
\includegraphics[width = 17.5cm]{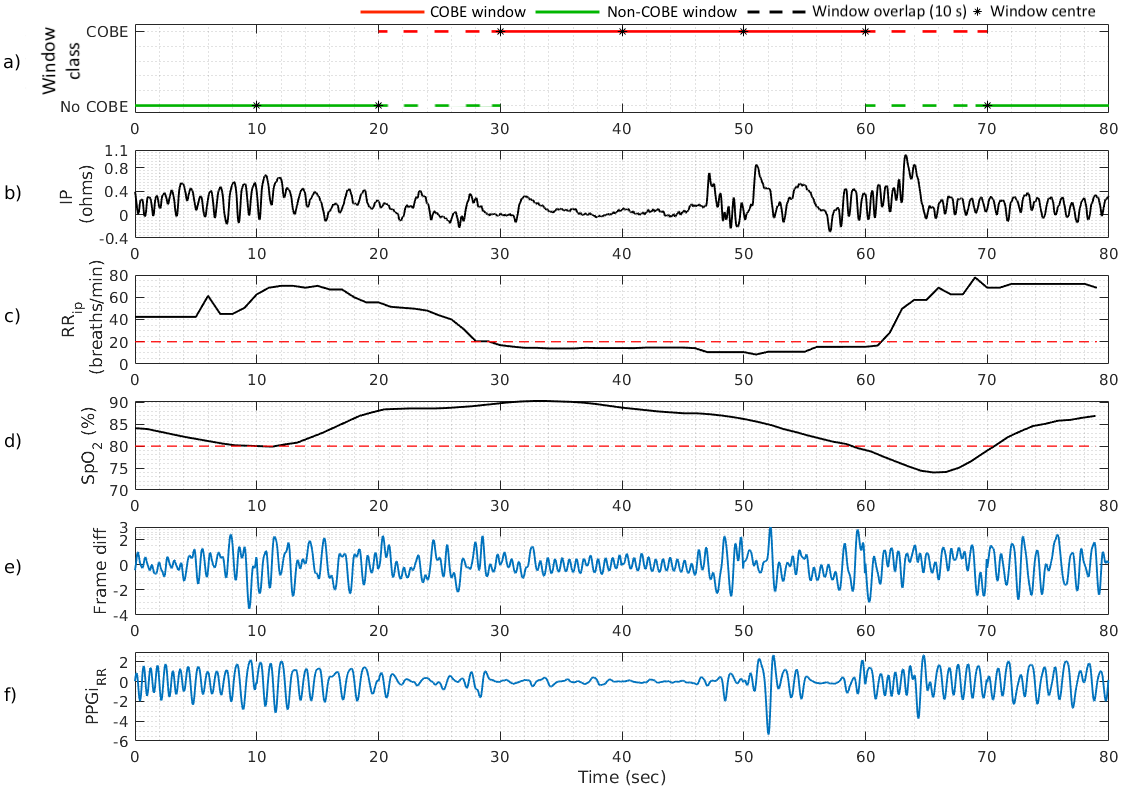}
\caption{Camera-derived respiratory signals extracted from a neonatal video recording, showing an example segment containing periods of normal breathing and a COBE episode. a) Overlapping 20-second analysis windows assigned binary class labels of COBE (red) or non-COBE (green). Black asterisks indicate the centres of consecutively labelled windows (10-s step size), while the coloured horizontal lines show the temporal extent of the corresponding analysis windows. Dashed line segments indicate the 10-s overlap between adjacent analysis windows. b) filtered IP signal; c) $RR_{ip}$ signal showing a respiratory pause. The red dashed line indicates the 20 breaths/min threshold used during COBE annotation; d) $SpO_{2}$ signal showing a period of desaturation. The red dashed line indicates the 80\% threshold used during COBE annotation; e) FD signal; f) $PPGi_{rr}$ signal.}
\label{cam_sig_extracted}
\end{figure}

The 80-second segments were then subdivided into overlapping 20-second windows with a 10-second step size, producing seven windows per 80-second segment. A 20-second window was sufficiently long to fully contain the shortest clinically relevant COBE episodes (10 seconds), while the 10-second step provided 50\% overlap between adjacent windows, allowing events occurring near window boundaries to be captured within at least one analysis window. Each window was assigned a binary class label (COBE or non-COBE) based on the expert annotations, as illustrated in figure~\ref{cam_sig_extracted}a. This procedure produced 4,823 windows, including 755 COBE and 4,068 non-COBE samples. Table~\ref{set_summary} summarises the final machine learning dataset. 

\begin{table}[!h]
\centering
\caption{Composition of the final annotated dataset before partitioning into training and test sets. The table summarises the number of annotated 80-second segments, derived 20-second analysis windows, and contributing infants.}
\label{set_summary}
\scalebox{1.0}{
\begin{tabular}{>{\raggedright\arraybackslash}p{5.4cm}
                R{3.5cm}}
\multicolumn{1}{c}{\textbf{Dataset component}} &
\multicolumn{1}{c}{\textbf{Count}} \\
\hline
Contributing infants & 23 \\
COBE segments (80 s) & 246 \\
Normal breathing segments (80 s) & 443 \\
\textbf{Total 80-second segments} & \textbf{689} \\
\hline
Positive 20-second windows & 755 \\
Negative 20-second windows & 4,068 \\
\textbf{Total analysis windows} & \textbf{4,823} \\
\hline
\end{tabular}}
\end{table}

\newpage
\subsubsection{Data splitting}
Each 20-second window was treated as an individual training instance. To prevent data leakage, infant-level separation was preserved throughout dataset partitioning such that windows from the same infant were not present in both training and test sets. The dataset of 23 infants was divided into a training set (19 infants) and an independent test set (4 infants). The training set was further partitioned into five cross-validation folds, with each fold containing data from 3--4 infants. Dataset partitions were selected to maintain demographic balance with respect to gestational age, sex, and the distribution of COBE and non-COBE windows.

\begin{table}[h!]
\centering
\caption[Training and test set window distribution.]{Distribution of positive and negative windows in the training and independent test sets.}
\label{set_split}
\scalebox{0.9}{
\begin{tabular}{|>{\centering\arraybackslash}p{3.0cm}|>{\centering\arraybackslash}p{3.0cm}|>{\centering\arraybackslash}p{3.0cm}|>{\centering\arraybackslash}p{3.0cm}|}
\hline
Dataset & Positive windows & Negative windows & Total windows \\
\hline
Training & 639 & 3,456 & 4,095\\
\hline
Test & 116 & 612 & 728 \\
\hline
Total & 755 & 4,068 & 4,823 \\
\hline
\end{tabular}}
\end{table}

\subsection{Machine learning approaches for COBE episode detection}
\begin{figure}[h!]
\centering
\includegraphics[width = 11.5cm]{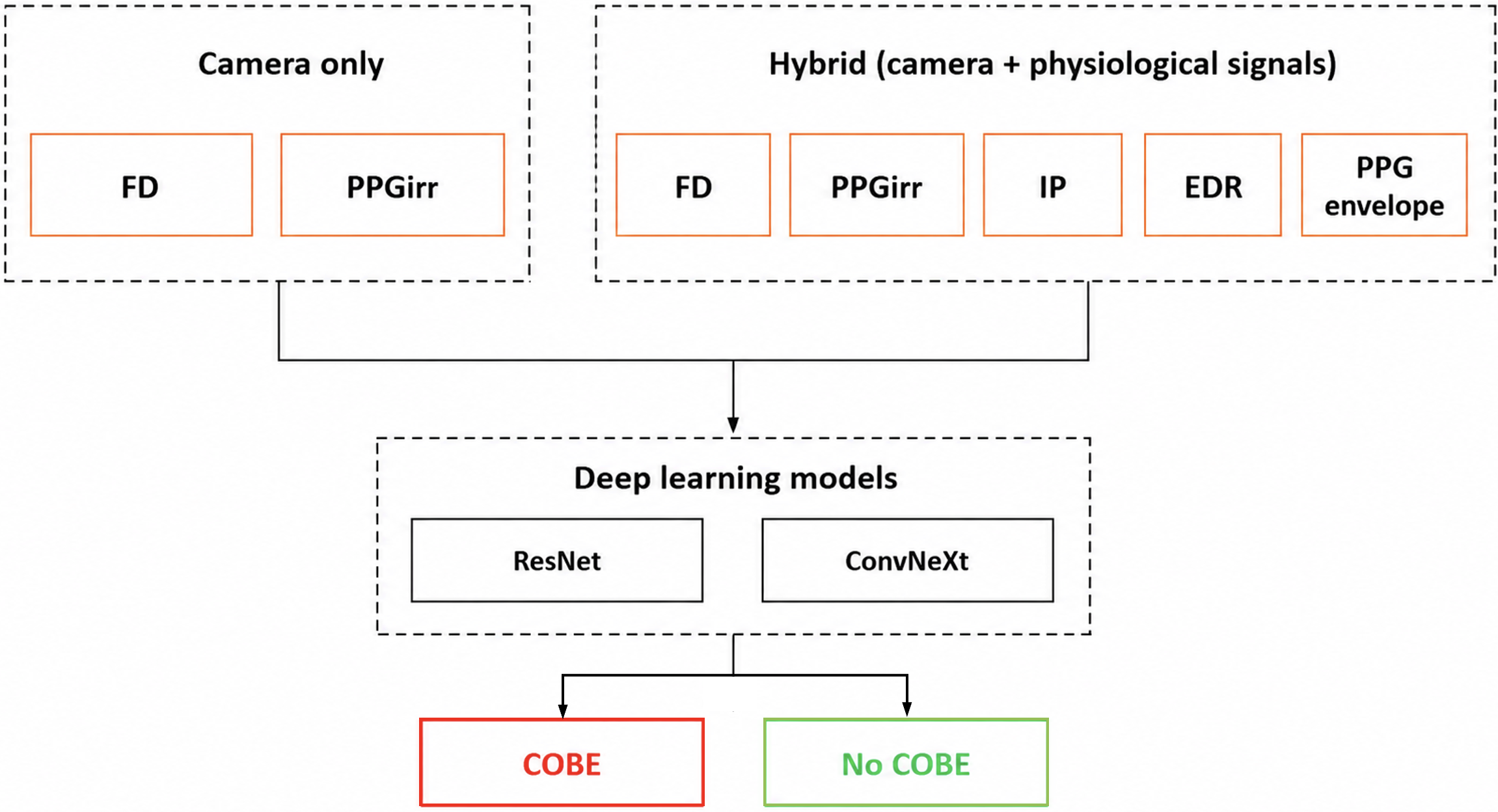}
\caption{Overview of the machine learning approaches evaluated for COBE detection. Camera-only models were trained using FD and $PPGi_{rr}$ signals.  Multimodal camera models combined FD and $PPGi_{rr}$, while hybrid models integrated camera-derived and physiological signals. ResNet and ConvNeXt architectures were evaluated for binary classification of COBE and non-COBE.}
\label{ml_overview}
\end{figure}

Figure~\ref{ml_overview} provides an overview of the machine learning approaches evaluated in this study. Residual Networks (ResNets) and ConvNeXt were investigated as the deep learning architectures for COBE detection. Both architectures were evaluated using three input configurations. Models were trained using individual camera-derived signals, combined camera-derived signals, and hybrid combinations of camera-derived and physiological signals. First, camera-only models were trained using the FD and $PPGi_{rr}$ signals extracted from neonatal video recordings. Second, multimodal camera models combined FD and $PPGi_{rr}$ using a late-fusion strategy. Finally, hybrid models integrated the camera-derived signals with physiological signals (IP, ECG-derived respiration (EDR), and the PPG envelope) using late fusion. While physiological signals provide high temporal resolution, they are susceptible to motion artefacts and electrode displacement. Conversely, camera-derived signals are non-contact but sensitive to occlusion and lighting variability. The hybrid framework was therefore designed to leverage complementary strengths across modalities and assess whether their integration improved COBE detection performance.

\paragraph{}
For multimodal and hybrid experiments, late fusion was employed. Each input modality was processed by an independent feature-extraction branch before the resulting feature representations were concatenated for classification. Although modalities were processed separately during feature extraction, all branches were trained jointly in an end-to-end manner using a shared classification objective, allowing gradients to optimise modality-specific feature representations simultaneously.

\paragraph{}
All models were trained and evaluated using the same dataset partitioning, windowing strategy, and evaluation protocol. Camera-only models were first evaluated using individual and combined camera-derived signals, after which the best-performing camera configurations were integrated with physiological signals in the hybrid framework.

\subsection{Camera-based COBE detection using ResNet}
\label{res_cam}
To evaluate the feasibility of camera-only COBE detection, ResNet architectures were adapted for one-dimensional time-series analysis. All two-dimensional convolutional, pooling, and normalisation layers were replaced with their one-dimensional counterparts to process the two extracted camera-derived signals (FD and $PPGi_{rr}$).

\subsubsection{Network architecture}
\label{res_archite}
\begin{figure}[h!]
\centering
\includegraphics[width = 15.5 cm]{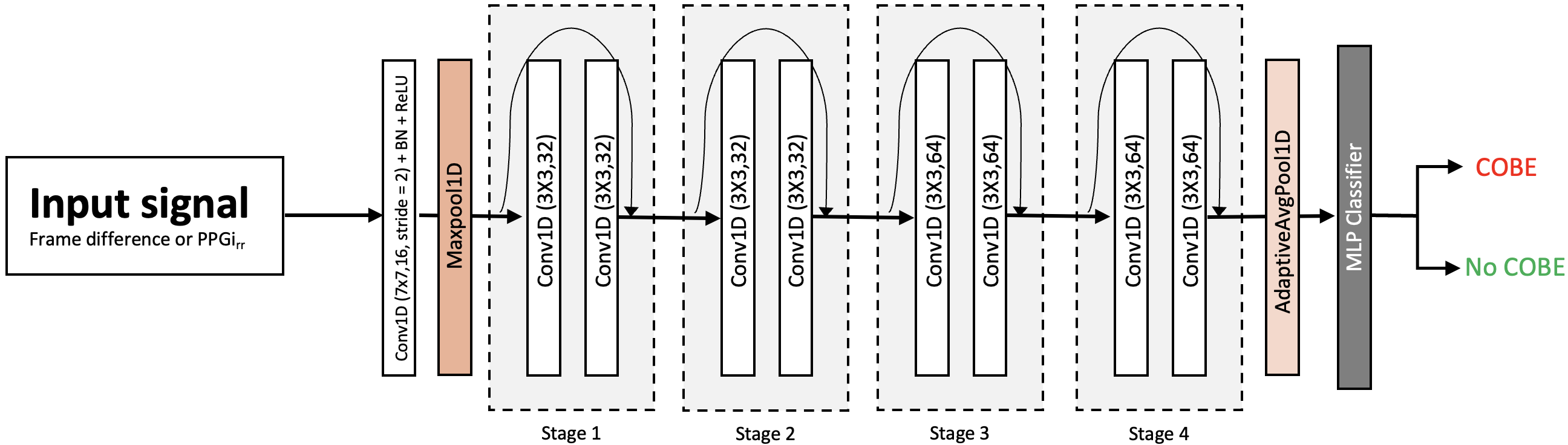}
\caption{Adapted one-dimensional ResNet architecture for camera-derived COBE detection. The network processes frame-difference (FD) and pixel-intensity ($PPGi_{rr}$) signals to produce a binary classification output.}
\label{resnet_designed}
\end{figure}

Following Carter \textit{et al.}~\cite{carter2023deep}, the number of feature channels per stage was reduced from the conventional [64,128,256,512] to [32,32,64,64], reflecting the lower feature complexity of one-dimensional signals compared with two-dimensional images. Adaptive average pooling was applied prior to classification to retain information from different regions of the feature map while maintaining a fixed-dimensional representation suitable for the classifier. The final classifier consisted of a multi-layer perceptron (MLP) replacing the standard single fully connected layer, enabling modelling of non-linear temporal feature interactions. Residual skip connections were retained throughout to maintain stable gradient propagation. The resulting architecture for camera-based COBE detection is shown in figure~\ref{resnet_designed}.

\subsubsection{Network training}
\label{res_train}
Three standard ResNet variants (ResNet-18, ResNet-34, and ResNet-50) were evaluated to assess the influence of network depth on camera-based COBE detection performance. Models were trained using the Adam optimiser with learning rates selected within the range $9 \times 10^{-5}$ to $1 \times 10^{-4}$ and exponential decay applied per epoch.  Five-fold cross-validation was performed on the training set for model selection. 

\subsubsection{Multimodal camera fusion}
\begin{figure}[h!]
\centering
\includegraphics[width = 13.5 cm]{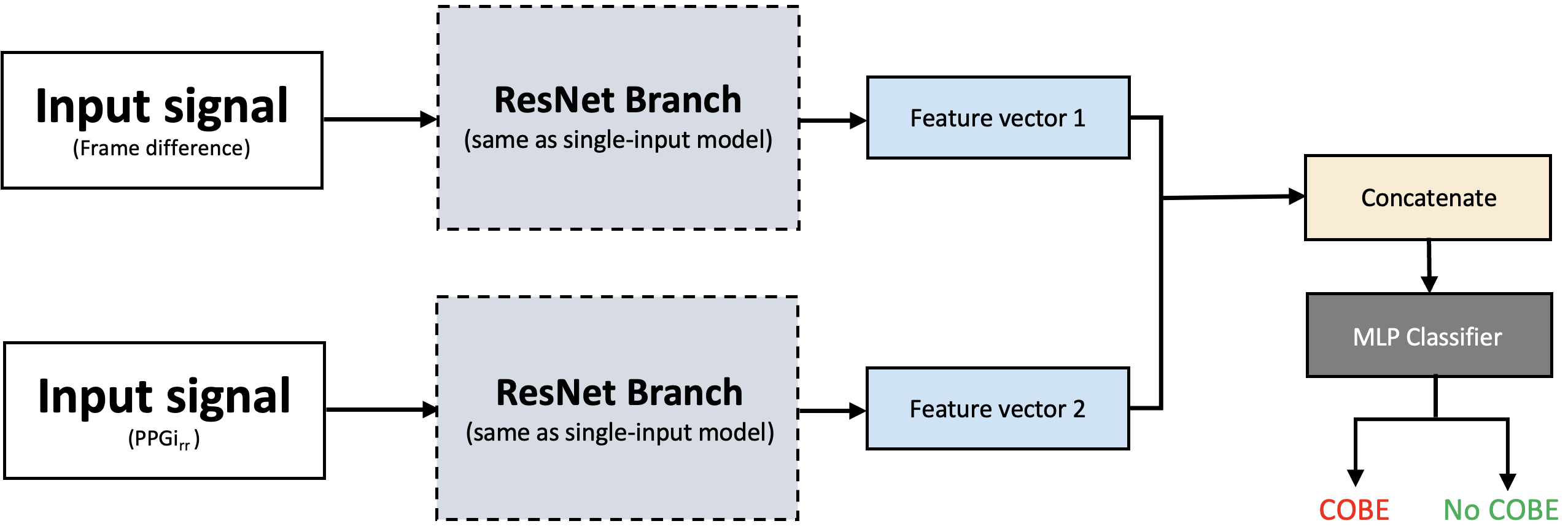}
\caption{Late-fusion ResNet architecture for multimodal camera-based COBE detection. Frame-difference (FD) and $PPGi_{rr}$ signals were processed independently using separate ResNet branches identical to the single-input architecture shown in figure~\ref{resnet_designed}. The resulting feature vectors were concatenated and passed to a shared multi-layer perceptron (MLP) classifier to predict COBE and non-COBE events.}
\label{resnet_multicam}
\end{figure}

To evaluate whether combining camera-derived motion and respiratory information improved COBE detection, FD and $PPGi_{rr}$ were also evaluated jointly using a late-fusion strategy. Each signal was processed by an independent ResNet branch, and the resulting feature representations were concatenated prior to classification as shown in figure~\ref{resnet_multicam}. This configuration enabled comparison between models trained on FD alone, $PPGi_{rr}$ alone, and the combined camera-derived signals.

\subsection{Hybrid COBE detection using camera and physiological signals}
Physiological respiratory signals were obtained from IP, ECG, and PPG recordings using the signal-processing pipeline described in Appendix~\ref{appendix_physio}. The filtered IP waveform was used directly as an input respiratory signal. An ECG-derived respiration (EDR) signal was computed from respiratory-induced variations in successive ECG R-peak amplitudes. Spline interpolation was then applied to obtain a continuous respiratory waveform. A respiratory envelope was derived from the PPG signal using peak detection followed by cubic spline interpolation. These approaches are consistent with established methods for extracting respiratory information from ECG and PPG signals~\cite{ponsiglione2025comparison,zarei2020automatic,charlton2017breathing}. The resulting respiratory signals were resampled to a common frequency of 60 Hz to facilitate multimodal analysis. 

\subsubsection{Hybrid architecture}
\begin{figure}[h!]
\centering
\includegraphics[width = 13.0cm]{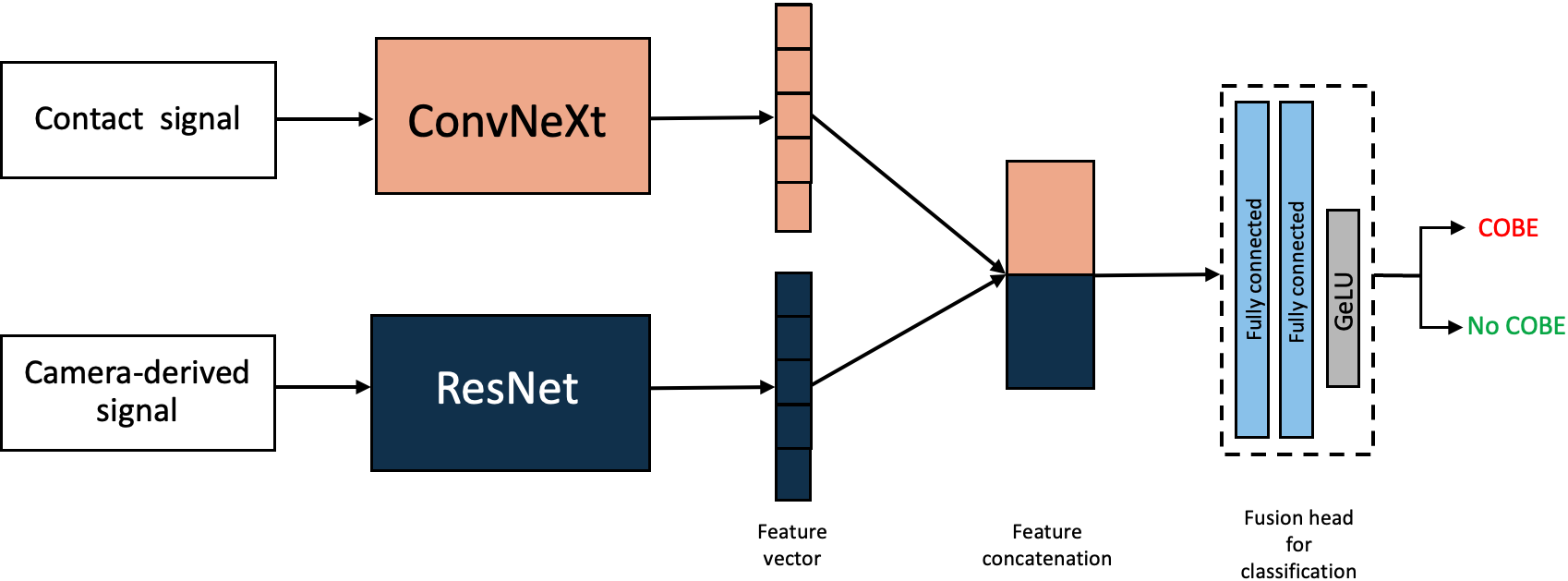}
\caption{Hybrid multimodal architecture for COBE detection. Camera-derived features extracted using ResNet were fused with physiological features extracted using ConvNeXt. The resulting feature vectors were concatenated and processed by a fusion head before final classification.}
\label{hybrid_arch}
\end{figure}

The hybrid network consisted of two modality-specific branches. The physiological branch employed a ConvNeXt architecture~\cite{Liu_2022_CVPR}, which extends conventional convolutional networks through the use of grouped convolutions, Gaussian Error Linear Unit (GELU) activations, and inverted bottleneck blocks. The camera branch employed the adapted ResNet architecture described in Section~\ref{res_cam}. Each branch independently extracted modality-specific feature representations, which were subsequently concatenated using a late-fusion strategy. Late fusion was selected to preserve modality-specific feature extraction while enabling joint end-to-end optimisation of both branches. The fusion head comprised two fully connected layers with GELU activations, enabling modelling of non-linear interactions between modalities. The final layer output the probability of a COBE episode for each analysis window. The overall hybrid architecture is illustrated in Figure~\ref{hybrid_arch}.

\subsubsection{Network training}
Both branches were trained jointly in an end-to-end manner using a shared weighted cross-entropy (CE) loss objective, defined as

\begin{equation}
\mathcal{L}_{\mathrm{CE}}
=
-\frac{1}{n}
\sum_{i=1}^{n}
\sum_{c=1}^{C}
w_c\,y_{i,c}\log(\hat{p}_{i,c}),
\end{equation}

where $n$ is the batch size, $C=2$ is the number of classes, $w_c$ is the weight assigned to class $c$, $y_{i,c}$ is the one-hot encoded ground-truth label for sample $i$, and $\hat{p}_{i,c}$ is the predicted probability that sample $i$ belongs to class $c$. The ConvNeXt branch was optimised using AdamW with a weight decay of 0.05, label smoothing of 0.1, exponential moving averaging of model weights, and a cosine learning-rate schedule with a five-epoch warm-up period. The ResNet branch was optimised using Adam. Gradients from the fusion head were propagated through both branches during training, enabling learning of complementary modality-specific representations. 

\subsection{Training and evaluation protocol}
Model development was performed using five-fold cross-validation on the training set. Following cross-validation, the best-performing configuration was retrained using the complete training set and evaluated on the independent test set.  Class imbalance was addressed using class-weighted cross-entropy loss, with higher loss weights assigned to the minority (COBE) class during training. Early stopping was applied when validation loss failed to improve for five consecutive epochs, preventing unnecessary training once performance had stabilised. Performance was evaluated using balanced accuracy, true positive rate (TPR), false positive rate (FPR), precision, F1 score, and Cohen's $\kappa$. Balanced accuracy was used as the primary metric to account for class imbalance and was calculated as:

\begin{equation}
\label{bal_accuracy}
\text{Balanced Accuracy}
=
\frac{\text{TPR}_{\mathrm{COBE}}+\text{TPR}_{\mathrm{Normal}}}{2}
\times 100\%
\end{equation}

Cross-validation results are reported as mean $\pm$ standard deviation across folds. Model selection prioritised configurations achieving the highest mean cross-validation balanced accuracy while favouring lower variance when performance was comparable.

\section{Results}
We evaluated the performance of camera-only ResNet models and hybrid multimodal models combining camera-derived and physiological signals. Results are presented for cross-validation as mean $\pm$ standard deviation, followed by evaluation on an independent test set.  

\paragraph{}
Cross-validation results for the ResNet models trained on individual camera-derived signals are summarised in Table~\ref{resnet_cv}. Among the single-signal models, the highest balanced accuracies were 63.3\% $\pm$ 4.3 for ResNet-50 trained on the FD signal and 68.7\% $\pm$ 5.8 for ResNet-34 trained on the $PPGi_{rr}$ signal. The multimodal model combining FD and $PPGi_{rr}$ achieved a balanced accuracy of 68.6\% $\pm$ 3.6.

\begin{table}[h!]
\centering
\caption{Five-fold cross-validation performance of ResNet models trained on camera-derived signals (FD, $PPGi_{rr}$, and their combination). Results reported as mean $\pm$ SD.}
\label{resnet_cv}
\scalebox{0.82}{
\begin{tabular}{|>{\raggedright\arraybackslash}p{2.2cm}|>{\raggedright\arraybackslash}p{1.2cm}|R{1.9cm}|R{1.9cm}|R{1.9cm}|R{1.9cm}|R{1.9cm}|R{2cm}|R{2cm}|R{2.3cm}|}
\hline
\multirow{4}{*}{\parbox[c]{2.0cm}{\centering Signal(s)}}&\multirow{4}{*}{\parbox[c]{1.2cm}{\centering Resnet Depth}} & \multicolumn{8}{c|}{Performance Metrics} \\
\cline{3-10}

& & \multicolumn{1}{c|}{TPR$\uparrow$} & \multicolumn{1}{c|}{FPR$\downarrow$} & \multicolumn{1}{c|}{Precision} &
\multicolumn{1}{c|}{F1 Score} & \multicolumn{1}{c|}{\parbox{1.3cm}{\centering Cohen Kappa}} & \multicolumn{1}{c|}{\parbox{2cm}{\centering No COBE\\Accuracy (\%)}} & \multicolumn{1}{c|}{\parbox{2cm}{\centering COBE\\Accuracy (\%)}} & \multicolumn{1}{c|}{\parbox{2cm}{\centering Balanced\\Accuracy (\%)}} \\

 \hline   
 \multirow{3}{*}{\parbox{0.7cm}{$FD$ }} & 18  & 0.57 $\pm$ 0.17 & 0.32 $\pm$ 0.10 & 0.27 $\pm$ 0.08 & 0.36 $\pm$ 0.09 & 0.18 $\pm$ 0.07 & 68.6 $\pm$ 10.4 & 57.1 $\pm$ 17.4 & 62.8 $\pm$ 4.9 \\

\cline{2-10}
& 34  & 0.67 $\pm$ 0.10 & 0.41 $\pm$ 0.07 & 0.25 $\pm$ 0.05 & 0.36 $\pm$ 0.06 & 0.15 $\pm$ 0.03 & 59.0 $\pm$ 6.8 & 66.7 $\pm$ 9.9 & 62.8 $\pm$ 2.5 \\

\cline{2-10}
& \textbf{50} & \textbf{0.60 $\pm$ 0.12} & \textbf{0.33 $\pm$ 0.07} & \textbf{0.27 $\pm$ 0.08} & \textbf{0.36 $\pm$ 0.08} & \textbf{0.18 $\pm$ 0.07} & \textbf{66.5 $\pm$ 7.2} & \textbf{60.1 $\pm$ 12.3} & \textbf{63.3 $\pm$ 4.3} \\

\hline
\hline   
\multirow{3}{*}{\parbox{1.0cm}{$PPGi_{rr}$}} & 18 & 0.77 $\pm$ 0.09 & 0.38 $\pm$ 0.06 & 0.29 $\pm$ 0.07 & 0.41 $\pm$ 0.08 & 0.23 $\pm$ 0.07 & 61.7 $\pm$ 6.2 & 76.7 $\pm$ 9.3 & 69.2 $\pm$ 5.2 \\
\cline{2-10}
& \textbf{34} & \textbf{0.71 $\pm$ 0.14} & \textbf{0.34 $\pm$ 0.07} & \textbf{0.30 $\pm$ 0.07} & \textbf{0.42 $\pm$ 0.08} & \textbf{0.24 $\pm$ 0.07} & \textbf{66.3 $\pm$ 6.9} & \textbf{71.0 $\pm$ 14.4} & \textbf{68.7 $\pm$ 5.8} \\
\cline{2-10}
& 50 & 0.70 $\pm$ 0.04 & 0.33 $\pm$ 0.07 & 0.30 $\pm$ 0.09 & 0.41 $\pm$ 0.08 & 0.24 $\pm$ 0.07 & 67.5 $\pm$ 6.8 & 69.6 $\pm$ 4.3 & 68.5 $\pm$ 2.8 \\
\hline
\hline  
$FD$ + $PPGi_{rr}$ &  & 0.76 $\pm$ 0.07 & 0.39 $\pm$ 0.05 & 0.28 $\pm$ 0.07 & 0.41 $\pm$ 0.07 & 0.22 $\pm$ 0.06 & 61.4 $\pm$ 4.8 & 75.8 $\pm$ 6.8 & 68.6 $\pm$ 3.6 \\
\hline   
\multicolumn{4}{l}{\footnotesize FD represent the frame difference signal} \\
\end{tabular}}
\end{table}

Table~\ref{resnet_test} reports performance of the best ResNet variants on the independent test set. The $PPGi_{rr}$-based ResNet-34 achieved balanced accuracy of 76.9\% on the independent test set (TPR = 0.80).  

\begin{table}[h!]
\centering
\caption{Test set performance of the best-performing models identified during five-fold cross-validation (bolded in Table~\ref{resnet_cv}). }
\label{resnet_test}
\scalebox{0.93}{
\begin{tabular}{|>{\raggedright\arraybackslash}p{2.5cm}|R{1.5cm}|R{1.5cm}|R{1.5cm}|R{1.5cm}|R{1.9cm}|R{2cm}|R{2cm}|R{2.1cm}|}
\hline
\multirow{4}{*}{\parbox[c]{2.5cm}{\centering Signal(s)}}& \multicolumn{8}{c|}{Performance Metrics} \\
\cline{3-9}
&  \multicolumn{1}{c|}{TPR$\uparrow$} & \multicolumn{1}{c|}{FPR$\downarrow$} & \multicolumn{1}{c|}{Precision} &
\multicolumn{1}{c|}{F1 Score} & \multicolumn{1}{c|}{\parbox{1.3cm}{\centering Cohen Kappa}} & \multicolumn{1}{c|}{\parbox{2cm}{\centering No COBE\\Accuracy (\%)}} & \multicolumn{1}{c|}{\parbox{2cm}{\centering COBE\\Accuracy (\%)}} & \multicolumn{1}{c|}{\parbox{2cm}{\centering Balanced\\Accuracy (\%)}} \\

\hline
$FD$ & 0.52 & 0.24 & 0.31 & 0.39 & 0.22 & 76.5 & 51.7 & 64.1 \\
\hline
$PPGi_{rr}$ & 0.80 & 0.26 & 0.38 & 0.52 & 0.37 & 73.6 & 80.2 & 76.9 \\
\hline
$FD$ + $PPGi_{rr}$ & 0.70 & 0.24 & 0.37 & 0.49 & 0.34 & 75.9 & 69.8 & 72.9 \\
\hline
\multicolumn{4}{l}{\footnotesize FD represent the frame difference signal} \\
\end{tabular}}
\end{table} 

Table~\ref{hybrid_table} summarises the performance of the hybrid models evaluated on the independent test set across different input combinations. Each configuration included at least one camera-derived signal combined with physiological information from IP, the PPG envelope,  EDR, or their combination. Extended hybrid results, including additional signal combinations and cross-validation outcomes, are provided in the supplementary material (Tables~\ref{hybrid_cvtable} - \ref{hybrid_testtable}).

\begin{table}[h!]
\centering
\caption{Summary of final hybrid model performance on the unseen test set. Each row represents a combination of at least one camera-derived signal and at least one physiological signal.}
\label{hybrid_table}
\scalebox{0.87}{
\begin{tabular}{|>{\raggedright\arraybackslash}p{3.5cm}|R{1.5cm}|R{1.5cm}|R{1.5cm}|R{1.5cm}|R{1.9cm}|R{2cm}|R{2cm}|R{2.3cm}|}
\hline
\multirow{4}{*}{\parbox[c]{3.5cm}{\centering Signal(s)}} & \multicolumn{8}{c|}{Performance Metrics} \\
\cline{2-9}
& \multicolumn{1}{c|}{TPR$\uparrow$} & \multicolumn{1}{c|}{FPR$\downarrow$} & \multicolumn{1}{c|}{Precision} &
\multicolumn{1}{c|}{F1 Score} & \multicolumn{1}{c|}{\parbox{1.3cm}{\centering Cohen Kappa}} & \multicolumn{1}{c|}{\parbox{2cm}{\centering No COBE\\Accuracy (\%)}} & \multicolumn{1}{c|}{\parbox{2cm}{\centering COBE\\Accuracy (\%)}} & \multicolumn{1}{c|}{\parbox{2cm}{\centering Balanced\\Accuracy (\%)}} \\
\hline
\multicolumn{9}{c}{\textbf{\textcolor{orange}{One camera + one physiological signal}}} \\ 
\hline
$FD$ + IP            & 0.81 & 0.08 & 0.68 & 0.74 & 0.68 & 92.1 & 81.0 & 86.6 \\
\hline
$FD$ + $PPG$  & 0.78 & 0.41 & 0.28 & 0.41 & 0.22 & 59.1 & 78.5 & 68.8\\
\hline
$FD$ + $EDR$  & 0.89 & 0.32 & 0.36 & 0.51 & 0.36 & 67.7 & 88.8 & 78.2 \\
\hline
\hline
$PPGi_{rr}$ + IP & \textbf{0.92} & \textbf{0.11} &\textbf{ 0.63} & \textbf{0.75} & \textbf{0.68} & \textbf{88.9} & \textbf{92.2} & \textbf{90.6} \\
\hline
$PPGi_{rr}$ + $PPG$       & 0.76 & 0.22 & 0.42 & 0.54 & 0.41 & 78.4 & 75.9 & 77.1 \\
\hline
$PPGi_{rr}$ + $EDR$  & 0.85 & 0.37 & 0.32 & 0.47 & 0.29 & 63.3 & 85.3 & 74.3 \\
\hline
\multicolumn{9}{c}{\textbf{\textcolor{orange}{Two cameras + one physiological signal}}} \\ 
\hline
$FD$ + $PPGi_{rr}$ + IP    & 0.91 & 0.12 & 0.60 & 0.72 & 0.65 & 87.7 & 90.5 & 89.1 \\
\hline
$FD$ + $PPGi_{rr}$ + $PPG$ &  0.80 & 0.29 & 0.36 & 0.50 & 0.34 & 71.0 & 80.2 & 75.6 \\
\hline
$FD$ + $PPGi_{rr}$ + $EDR$ & 0.72 &  0.19 &  0.44 & 0.55 & 0.42 & 81.0 & 72.4 & 76.7 \\
\hline
\multicolumn{9}{c}{\textbf{\textcolor{orange}{Two cameras + three physiological signals}}} \\ 
\hline
$FD$ + $PPGi_{rr}$ + IP + PPG + EDR  &  \multirow{2}{*}{0.76} &  \multirow{2}{*}{0.08} &  \multirow{2}{*}{0.66} &  \multirow{2}{*}{0.71} &  \multirow{2}{*}{0.64} &  \multirow{2}{*}{92.1} &  \multirow{2}{*}{75.9} &  \multirow{2}{*}{84.0} \\
\hline
\multicolumn{2}{l}{\footnotesize FD represent the frame difference signal} \\

\end{tabular}}
\end{table}

\section{Discussion}
\label{discuss}
This study evaluated the feasibility of detecting apnoea-related cessation of breathing (COBE) in pre-term infants using video camera-derived signals, both independently and in combination with physiological measurements. Following video quality assessment, 689 annotated 80-second segments from 23 infants were retained for analysis.  The results demonstrate that non-contact video-derived features contain clinically relevant information for distinguishing COBE episodes from normal breathing, supporting the feasibility of video-based COBE detection in the NICU. 

\paragraph{}
Camera-only models demonstrated that respiratory motion extracted from video can be used to detect COBE events. Among the camera-derived signals, the pixel-intensity respiratory signal ($PPGi_{rr}$) consistently outperformed the frame-difference (FD) signal, achieving a test accuracy of 76.9\% compared to 64.1\% for FD. While FD captures global torso motion, including non-respiratory movement, $PPGi_{rr}$ focuses on localised abdominal motion and more directly reflects breathing-related dynamics. This suggests that respiration-specific motion features provide more discriminative information than general movement alone. These findings support the feasibility of non-contact respiratory monitoring in neonatal environments. 

\paragraph{}
Combining FD and $PPGi_{rr}$ did not improve performance beyond $PPGi_{rr}$ alone (72.9\% vs 76.9\%), indicating that global motion information may introduce variability that obscures respiration-specific patterns. This may reflect the difficulty of distinguishing respiratory from non-respiratory motion in smaller datasets, highlighting the importance of feature selection in multimodal settings, particularly when training data are limited.

\paragraph{}
The strongest performance was achieved using hybrid models integrating video camera-derived signals with impedance pneumography (IP). The combination of $PPGi_{rr}$ and IP achieved a test accuracy of 90.6\%, exceeding the performance of video camera-only models and indicating a clear benefit of multimodal integration. For comparison, contact-only machine learning models evaluated on the same dataset achieved a maximum test accuracy of 88.7\%~\cite{serame2026contact}. The hybrid configuration therefore achieved comparable and slightly higher performance while incorporating visual information. Non-contact sensing can provide complementary respiratory information when combined with conventional physiological monitoring, particularly IP. While IP provides a direct measure of respiratory effort, it remains susceptible to motion artefacts and electrode displacement. Camera-derived signals contribute contextual motion cues and additional respiratory dynamics that enhance classification performance when integrated within a unified model.

\paragraph{}
In contrast, combining camera-derived signals with PPG or ECG alone yielded more modest improvements. For example, combining $PPGi_{rr}$ with PPG resulted in only a marginal increase in performance compared to $PPGi_{rr}$ alone, suggesting overlap in the physiological information captured by these signals. Similarly, combining FD with ECG improved performance relative to models trained on either signal alone. Notably, indiscriminate fusion of all available signals did not further improve performance (84\%), indicating that adding redundant inputs may not introduce informative features. These observations emphasise that multimodal integration must be selective and physiologically motivated.

\subsection{Limitations}
A primary limitation of this study is the modest dataset size, reflecting the challenges of acquiring high-quality neonatal video recordings in clinical environments. Although segment quality assessment was necessary to ensure reliable motion extraction, it reduced the number of analysable COBE events and limits evaluation across a wider range of clinical conditions. Additionally, BlazePose was originally trained primarily on adult data. Although it enabled ROI tracking for most analysed recordings, dedicated neonatal pose-estimation models may further improve robustness.

\paragraph{}
The hybrid framework was evaluated retrospectively. Future work should assess performance under varying lighting conditions, occlusion patterns, and caregiving activities. Larger multi-centre datasets will be essential to evaluate generalisability across diverse neonatal populations and incubator configurations.

\section{Conclusion}
This study demonstrates the feasibility of video camera-based detection of COBE in pre-term infants and shows that video-derived respiratory features contain clinically meaningful information. Video camera-only models achieved moderate detection performance, demonstrating feasibility. Hybrid integration with impedance pneumography improved accuracy, supporting the complementary value of non-contact sensing. These findings indicate that video camera-based monitoring can augment conventional NICU respiratory monitoring by providing additional contextual and motion-related information. Rather than replacing contact-based sensors, multimodal fusion offers a strategy for enhancing detection robustness in dynamic neonatal environments.

\newpage
\appendix
\renewcommand{\thetable}{S\arabic{table}}
\setcounter{table}{0} 

\section{Supplemetary methods: Recomputing reference RR}
\label{appendix:rr_ip}
A reference respiratory rate ($RR_{ip}$) was recomputed directly from the raw impedance pneumography (IP) signal. The IP waveform was first resampled from 62.5 Hz to 24 Hz using cubic spline interpolation and de-trended to remove the DC component. An 8th-order high-pass Butterworth filter (0.033 Hz) and a 6th-order low-pass Butterworth filter (2.83 Hz) were then applied to remove baseline drift and high-frequency noise while preserving the physiological respiratory frequency range observed in the study population.

\paragraph{}
Respiratory cycles were identified using a peak-and-trough detection algorithm based on the Mean Average Curve (MAC) method described by Lu \textit{et al.}~\cite{lu2006semi}. The MAC signal provided a dynamic threshold for peak detection, with respiratory peaks defined as local maxima above the threshold and troughs identified as local minima between successive peaks. Peaks with amplitudes below 20\% of the median respiratory amplitude were rejected.

\paragraph{}
Signal quality assessment was performed for each detected respiratory cycle using the methodology proposed by Li \textit{et al.}~\cite{li2007robust}. Three complementary signal quality indices (SQIs) were computed. First, a physiological-bounding SQI ($SQI_{phys}$) assessed whether the instantaneous respiratory rate fell within a plausible range for pre-term infants (2--170 breaths/min). Second, a spectral-concentration SQI ($SQI_{bin}$) quantified the proportion of signal power concentrated around the dominant respiratory frequency. $SQI_{bin}$ was assigned a value of 1 when at least 50\% of the spectral power was contained within a 0.3 Hz bandwidth centred on the dominant frequency~\cite{chaichulee2018non}. Third, the Spectral Purity Index (SPI) quantified the extent to which the respiratory waveform was dominated by a single periodic frequency component, with lower values indicating increased waveform irregularity, noise, or motion artefacts.

\paragraph{}
The three measures were combined to produce a breath-level signal quality index ($SQI_{breath}$). Respiratory cycles failing any of the quality criteria were excluded from respiratory rate estimation. This quality-control procedure reduced the influence of motion artefacts, electrode disturbances, and non-respiratory fluctuations on the recomputed respiratory rate.

\paragraph{}
After detecting the peaks and troughs of the respiratory signal and assessing the signal quality for each breath, respiratory rate was estimated using a breath-counting approach within a 10-second sliding window with a step size of 1 second. The number of accepted respiratory cycles within each window was converted to breaths per minute by scaling to a 60-second interval, producing the continuous $RR_{ip}$ time series.

\newpage
\section{Supplemetary methods: Creating a COBE dataset}
\label{appendix:supp_dataset}
\begin{figure}[h!]
\centering
\includegraphics[width = 13.5 cm]{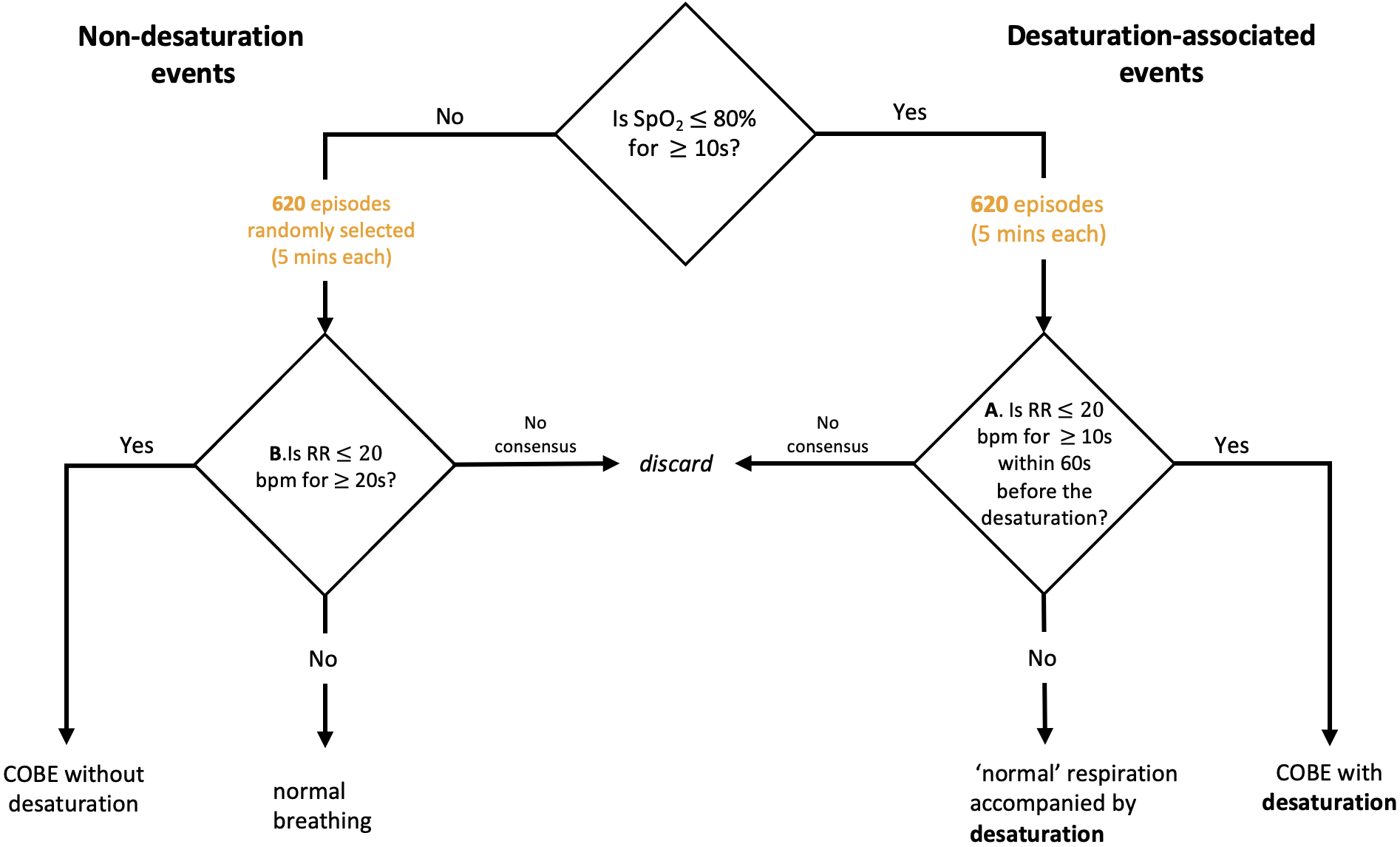}
 \caption{Review workflow used for reference labelling of COBE events. Annotators examined synchronised respiratory rate, oxygen saturation, and waveform signals within extended contextual windows to determine whether a true cessation of breathing was present.}
         \label{annotate_process}
    \end{figure}

\section{Supplementary methods: Camera-based respiratory rate estimation}
\label{appendix:rr_cam}
To support respiratory rate estimation during both normal breathing and COBE episodes, a motion signal was first computed from the torso ROI following the frame-difference approach described by Cattani \textit{et al.}~\cite{cattani2014wire}. The absolute pixel-wise difference between consecutive video frames was calculated within the torso ROI and averaged across pixels within the ROI to quantify infant motion.

\paragraph{}
A camera-derived respiratory rate ($RR_{cam}$) was then estimated from the respiratory signal ($PPGi_{rr}$) extracted from the abdominal ROI. The respiratory signal was obtained from the green colour channel, which has been shown to capture respiratory-related motion effectively in skin regions~\cite{chaichulee2018non,ahani2024video}.

\paragraph{}
The $PPGi_{rr}$ signal was de-trended and filtered using a motion-dependent filter-switching approach. Periods of normal activity and low activity associated with respiratory pauses were identified using the computed motion signal, and different filter sets were applied to each activity state. During normal activity, an 8th-order high-pass filter (0.42 Hz) and a 6th-order low-pass filter (2.75 Hz) were applied, corresponding to respiratory rates between 25 and 140 breaths/min. During periods of low activity, a 6th-order high-pass filter (0.20 Hz) and a 2nd-order low-pass filter (1.42 Hz) were applied, corresponding to respiratory rates between 12 and 85 breaths/min. 

\paragraph{}
Respiratory cycles were identified using the Mean Average Curve (MAC) peak detector developed by Lu \textit{et al.}~\cite{lu2006semi} and the Boxed Slope Sum Function (BSSF) peak detector described by Zong \textit{et al.}~\cite{zong2003open}. Signal quality assessment was performed using two complementary measures. First, an activity-based signal quality index ($SQI_{act}$) quantified the extent of motion artefacts using the frame-difference signal. Second, a peak-agreement signal quality index ($SQI_{peak}$) quantified agreement between respiratory cycles identified by the MAC detector and the BSSF. These measures were combined to produce a breath-level signal quality index ($SQI_{breath}$), and respiratory cycles failing the quality criteria were excluded from respiratory rate estimation.

\paragraph{}
Respiratory rate was estimated using a breath-counting approach within a 10-second sliding window with a 1-second step size. Each window was expanded to include the complete first and last respiratory cycles before breath counting was performed. The number of accepted respiratory cycles within the expanded window was converted to breaths per minute, producing the continuous $RR_{cam}$ time series. The resulting RR estimates were subsequently aligned with the reference respiratory rate ($RR_{ip}$) using cross-correlation prior to dataset construction and analysis.

\section{Supplementary methods: Physiological respiratory signal extraction}
\label{appendix_physio}
\subsection{Impedance pneumography (IP)}
The IP signal was filtered using an 8th-order high-pass Butterworth filter (0.08 Hz) and a 6th-order low-pass Butterworth filter (2.75 Hz) to remove baseline drift and high-frequency noise while preserving the neonatal respiratory frequency range. The filtered waveform was resampled to 60 Hz and used directly as a measure of respiratory activity.

\subsection{ECG-derived respiration (EDR)}
The ECG signal was filtered using an 8th-order high-pass Butterworth filter (0.67 Hz) and a 2nd-order low-pass Butterworth filter (4 Hz). R-peaks were identified using the Pan–Tompkins algorithm~\cite{pan1985real}. Respiratory information was obtained from variations in successive ECG R-peak amplitudes, which were interpolated using cubic splines to generate a continuous ECG-derived respiration (EDR) waveform. The resulting signal was resampled to 60 Hz.

\subsection{PPG-derived respiratory envelope}
The PPG signal was filtered using an 8th-order high-pass Butterworth filter (0.08 Hz) and a 2nd-order low-pass Butterworth filter (2.75 Hz). Successive PPG peaks were identified and interpolated using cubic splines to generate a continuous respiratory envelope representing respiratory-induced modulation of the PPG waveform. The resulting signal was resampled to 60 Hz.

\begin{figure}[h!]
\centering
\includegraphics[width=12cm]{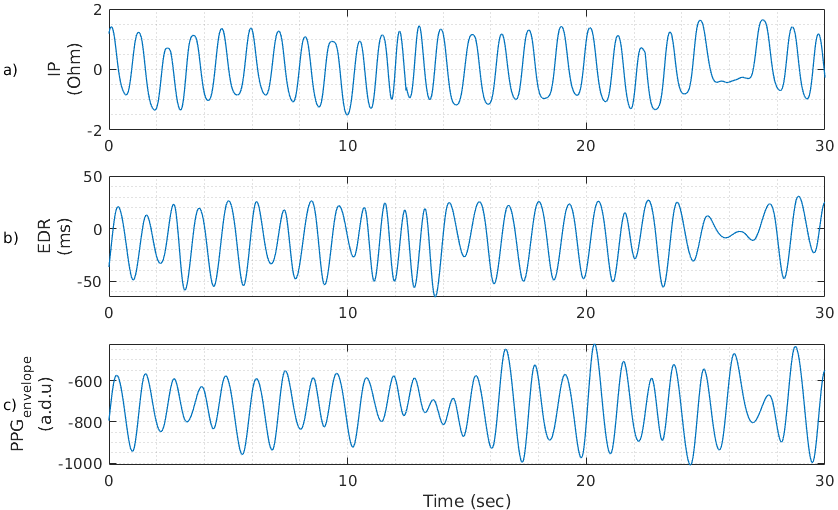}
\caption{Examples of respiratory signals extracted from synchronised physiological recordings and used for machine learning. a) IP waveform b) ECG-derived respiration (EDR),  and c) the PPG-derived respiratory envelope.}
\label{resp_signals}
\end{figure}

\section{COBE detection using a combination of camera  and physiological signals}
\label{appendix:suppHybridCOBE}
Tables~\ref{hybrid_cvtable} and~\ref{hybrid_testtable} show the performances of hybrid models under different input combinations. Each hybrid model combines camera-derived signals with physiological inputs (IP, $ECG$, and PPG envelope) to detect COBE episodes. 

\begin{table}[h!]
\centering
\caption{Cross-validation performance of hybrid models trained by combining camera-derived signals with physiological inputs. The best-performing combination is highlighted in bold.}
\label{hybrid_cvtable}
\scalebox{0.77}{
\begin{tabular}{|>{\raggedright\arraybackslash}p{3.5cm}|R{1.9cm}|R{1.9cm}|R{1.9cm}|R{1.9cm}|R{1.9cm}|R{2cm}|R{2cm}|R{2.3cm}|}
\hline
\multirow{4}{*}{\parbox[c]{3.5cm}{\centering Signal(s)}} & \multicolumn{8}{c|}{Performance Metrics} \\
\cline{2-9}
& \multicolumn{1}{c|}{TPR$\uparrow$} & \multicolumn{1}{c|}{FPR$\downarrow$} & \multicolumn{1}{c|}{Precision} &
\multicolumn{1}{c|}{F1 Score} & \multicolumn{1}{c|}{\parbox{1.3cm}{\centering Cohen Kappa}} & \multicolumn{1}{c|}{\parbox{2cm}{\centering No COBE\\Accuracy (\%)}} & \multicolumn{1}{c|}{\parbox{2cm}{\centering COBE\\Accuracy (\%)}} & \multicolumn{1}{c|}{\parbox{2cm}{\centering Balanced\\Accuracy (\%)}} \\
\hline
\multicolumn{9}{c}{\textbf{\textcolor{orange}{One camera + one physiological signals}}} \\ 
\hline
$FD$ + IP  & \textbf{0.86 $\pm$ 0.04} & \textbf{0.20 $\pm$ 0.02} & \textbf{0.46 $\pm$ 0.10} & \textbf{0.59 $\pm$ 0.09} & \textbf{0.48 $\pm$ 0.09} & \textbf{79.9 $\pm$ 2.4} & \textbf{85.4 $\pm$ 4.3} & \textbf{82.6 $\pm$ 2.6} \\
\hline
$FD$ + $PPG$ & 0.78 $\pm$ 0.10 & 0.43 $\pm$ 0.08 & 0.26 $\pm$ 0.04 & 0.39 $\pm$ 0.04 & 0.19 $\pm$ 0.06 & 56.6 $\pm$ 7.5 & 78.1 $\pm$ 9.7 & 67.4 $\pm$ 6.1 \\
\hline
$FD$ + $ECG$ & 0.78 $\pm$ 0.08 & 0.39 $\pm$ 0.08 & 0.29 $\pm$ 0.06 & 0.42 $\pm$ 0.06 & 0.23 $\pm$ 0.07 & 61.2 $\pm$ 8.1 & 78.1 $\pm$ 8.8 & 69.6 $\pm$ 5.0 \\
\hline
\hline
$PPGi_{rr}$ + IP &\textbf{0.81 $\pm$ 0.08} & \textbf{0.23 $\pm$ 0.05} & \textbf{0.42 $\pm$ 0.12} & \textbf{0.54 $\pm$ 0.11} & \textbf{0.42 $\pm$ 0.11} &\textbf{77.4 $\pm$ 4.5} & \textbf{81.0 $\pm$ 8.0} & \textbf{79.2 $\pm$ 3.9} \\
\hline
$PPGi_{rr}$ + $PPG$ & 0.67 $\pm$ 0.05 & 0.37 $\pm$ 0.07 & 0.27 $\pm$ 0.05 & 0.37 $\pm$ 0.05 & 0.18 $\pm$ 0.07 & 62.6 $\pm$ 6.8 & 67.1 $\pm$ 5.4 & 64.8 $\pm$ 5.4 \\
\hline
$PPGi_{rr}$ + $ECG$ & 0.68 $\pm$ 0.05 & 0.35 $\pm$ 0.06 & 0.28 $\pm$ 0.06 & 0.39 $\pm$ 0.06 & 0.21 $\pm$ 0.06 & 65.0 $\pm$ 6.2 & 68.3 $\pm$ 5.1 & 66.7 $\pm$ 3.5 \\
\hline
\hline
\multicolumn{9}{c}{\textbf{\textcolor{orange}{One camera + two physiological signals}}} \\ 
\hline
$FD$ + $PPG$ + IP & 0.73 $\pm$ 0.06 & 0.21 $\pm$ 0.03 & 0.40 $\pm$ 0.09 & 0.51 $\pm$ 0.08 & 0.39 $\pm$ 0.07 & 78.6 $\pm$ 2.7 & 73.0 $\pm$ 5.9 & 75.8 $\pm$ 2.4 \\
\hline
$FD$ + $ECG$ + IP & \textbf{0.88 $\pm$ 0.04} & \textbf{0.20 $\pm$ 0.03} & \textbf{0.47 $\pm$ 0.09} & \textbf{0.61 $\pm$ 0.08} & \textbf{0.50 $\pm$ 0.08} & \textbf{80.2 $\pm$ 2.8} & \textbf{87.8 $\pm$ 4.3} & \textbf{84.0 $\pm$ 2.0} \\
\hline
  $FD$ + $ECG$ + $PPG$  & 0.79 $\pm$ 0.08 & 0.39 $\pm$ 0.06 & 0.28 $\pm$ 0.05 & 0.42 $\pm$ 0.05 & 0.23 $\pm$ 0.05 & 60.8 $\pm$ 5.7 & 79.0 $\pm$ 7.6 & 69.9 $\pm$ 4.5 \\  
\hline
\hline
 $PPGi_{rr}$ + $PPG$ + IP & 0.80 $\pm$ 0.07 & 0.21 $\pm$ 0.03 & 0.43 $\pm$ 0.08 & 0.55 $\pm$ 0.08 & 0.43 $\pm$ 0.07 & 78.9 $\pm$ 3.1 & 79.7 $\pm$ 6.8 & 79.3 $\pm$ 2.4 \\
  \hline
 $PPGi_{rr}$ +$ECG$ + IP & \textbf{0.78 $\pm$ 0.08} & \textbf{0.21 $\pm$ 0.03} & \textbf{0.43 $\pm$ 0.10} & \textbf{0.55 $\pm$ 0.10} & \textbf{0.43 $\pm$ 0.10} & \textbf{79.3 $\pm$ 3.1} & \textbf{77.6 $\pm$ 7.5} & \textbf{78.5 $\pm$ 3.8} \\
  \hline
 $PPGi_{rr}$ +$ECG$ + $PPG$ & 0.71 $\pm$ 0.07 &  0.36 $\pm$ 0.05 & 0.28 $\pm$ 0.05 & 0.40 $\pm$ 0.05 &  0.21 $\pm$ 0.05 & 63.5 $\pm$ 5.2 & 71.0 $\pm$ 6.9 & 67.2 $\pm$ 3.8 \\
\hline
\hline
\multicolumn{9}{c}{\textbf{\textcolor{orange}{Two camera + one physiological signal}}} \\ 
\hline
$FD$ + $PPGi_{rr}$ + IP & 0.80 $\pm$ 0.08 & 0.23 $\pm$ 0.03 & 0.41 $\pm$ 0.11 & 0.54 $\pm$ 0.11 & 0.41 $\pm$ 0.12 & 77.0 $\pm$ 3.5 & 79.8 $\pm$ 7.5 & 78.4 $\pm$ 4.5 \\
\hline
$FD$ + $PPGi_{rr}$ + $PPG$ & 0.84 $\pm$ 0.08 & 0.42 $\pm$ 0.09 & 0.29 $\pm$ 0.06 & 0.43 $\pm$ 0.06 & 0.24 $\pm$ 0.08 & 58.6 $\pm$ 8.6 & 83.6 $\pm$ 7.8 & 71.1 $\pm$ 5.7 \\
\hline
$FD$ + $PPGi_{rr}$ + $ECG$ &  0.74 $\pm$ 0.05 & 0.38 $\pm$ 0.07 & 0.28 $\pm$ 0.06 & 0.40 $\pm$ 0.06 &  0.21 $\pm$ 0.06 & 62.1 $\pm$ 6.9 &  73.8 $\pm$ 4.9 &  68.0 $\pm$ 3.8 \\
\hline
\multicolumn{9}{c}{\textbf{\textcolor{orange}{Two camera + three physiological signals}}} \\ 
\hline
$FD$ + $PPGi_{rr}$ + IP + $PPG$ + $ECG$ & 0.90 $\pm$ 0.03 & 0.19 $\pm$ 0.02 & 0.48 $\pm$ 0.07 & 0.62 $\pm$ 0.07 & 0.52 $\pm$ 0.06 & 80.6 $\pm$ 2.1 & 90.2 $\pm$ 2.5 & 85.4 $\pm$ 1.7 \\
\hline
\multicolumn{2}{l}{\footnotesize FD represent the frame difference signal} \\

\end{tabular}}
\end{table}

\begin{table}[h!]
\centering
\caption[Test set performance of the hybrid models.]{Test set performance of the hybrid models, with the best models highlighted per camera signal. Each model was trained on a dataset of 19 infants and evaluated on the remaining 4 infants. Results reflect the final generalisation ability on held-out data.}
\label{hybrid_testtable}
\scalebox{0.83}{
\begin{tabular}{|>{\raggedright\arraybackslash}p{3.5cm}|R{1.9cm}|R{1.9cm}|R{1.9cm}|R{1.9cm}|R{1.9cm}|R{2cm}|R{2cm}|R{2.3cm}|}
\hline
\multirow{4}{*}{\parbox[c]{3.5cm}{\centering Signal(s)}} & \multicolumn{8}{c|}{Performance Metrics} \\
\cline{2-9}
& \multicolumn{1}{c|}{TPR$\uparrow$} & \multicolumn{1}{c|}{FPR$\downarrow$} & \multicolumn{1}{c|}{Precision} &
\multicolumn{1}{c|}{F1 Score} & \multicolumn{1}{c|}{\parbox{1.3cm}{\centering Cohen Kappa}} & \multicolumn{1}{c|}{\parbox{2cm}{\centering No COBE\\Accuracy (\%)}} & \multicolumn{1}{c|}{\parbox{2cm}{\centering COBE\\Accuracy (\%)}} & \multicolumn{1}{c|}{\parbox{2cm}{\centering Balanced\\Accuracy (\%)}} \\
\hline
\multicolumn{9}{c}{\textbf{\textcolor{orange}{One camera + one physiological signals}}} \\ 
\hline
$FD$ + IP            & \textbf{0.81} & \textbf{0.08} & \textbf{0.68} & \textbf{0.74} &\textbf{ 0.68} & \textbf{92.1} & \textbf{81.0} & \textbf{86.6} \\
\hline
$FD$ + $PPG$  & 0.78 & 0.41 & 0.28 & 0.41 & 0.22 & 59.1 & 78.5 & 68.8\\
\hline
$FD$ + $ECG$  & 0.89 & 0.32 & 0.36 & 0.51 & 0.36 & 67.7 & 88.8 & 78.2 \\
\hline
\hline
$PPGi_{rr}$ + IP & \textbf{0.92} & \textbf{0.11} &\textbf{ 0.63} & \textbf{0.75} & \textbf{0.7} & \textbf{88.9} & \textbf{92.2} & \textbf{90.6} \\
\hline
$PPGi_{rr}$ + $PPG$       & 0.76 & 0.22 & 0.42 & 0.54 & 0.41 & 78.4 & 75.9 & 77.1 \\
\hline
$PPGi_{rr}$ + $ECG$  & 0.85 & 0.37 & 0.32 & 0.47 & 0.29 & 63.3 & 85.3 & 74.3 \\
\hline
\hline
\multicolumn{9}{c}{\textbf{\textcolor{orange}{One camera + two physiological signals}}} \\ 
\hline
$FD$ + $PPG$ + IP & 0.85 & 0.07 & 0.71 & 0.78 & 0.73 & 93.0 & 85.3 & 89.2 \\
\hline
  $FD$ + $ECG$ + IP & \textbf{0.91} & \textbf{0.12} & \textbf{0.61} & \textbf{0.73} & \textbf{0.66} & \textbf{88.1} & \textbf{91.4} & \textbf{89.7} \\
\hline
  $FD$ + $ECG$ + $PPG$ & 0.85 & 0.32 & 0.35 & 0.50 & 0.34 & 68.2 & 85.3 & 76.8 \\
\hline
\hline
 $PPGi_{rr}$ + $PPG$ + IP & 0.91 & 0.13 & 0.58 & 0.71 & 0.63 & 86.8 & 90.5 & 88.7 \\
  \hline
 $PPGi_{rr}$ +$ECG$ + IP & \textbf{0.91} & \textbf{0.13} & \textbf{0.59} & \textbf{0.71} & \textbf{0.64} & \textbf{86.8} & \textbf{91.4} & \textbf{89.1} \\
  \hline
 $PPGi_{rr}$ +$ECG$ + $PPG$ & 0.88 & 0.25 & 0.41 & 0.56 & 0.43 & 74.5 & 87.9 & 81.2 \\
\hline
\hline
\multicolumn{9}{c}{\textbf{\textcolor{orange}{Two camera + one physiological signal}}} \\ 
\hline
$FD$ + $PPGi_{rr}$ + IP    & 0.91 & 0.12 & 0.60 & 0.72 & 0.65 & 87.7 & 90.5 & 89.1 \\
\hline
$FD$ + $PPGi_{rr}$ + $PPG$ & 0.80 & 0.29 & 0.36 & 0.50 & 0.34 & 71.0 & 80.2 & 75.6 \\
\hline
$FD$ + $PPGi_{rr}$ + $ECG$ & 0.72 & 0.19 & 0.44 & 0.55 & 0.42 & 81.0 & 72.4 & 76.7 \\
\hline
\multicolumn{9}{c}{\textbf{\textcolor{orange}{Two camera + three physiological signals}}} \\ 
\hline
$FD$ + $PPGi_{rr}$ + IP + $PPG$ + $ECG$  & 0.76 & 0.08 & 0.66 & 0.71 & 0.64 & 92.1 & 75.9 & 84.0 \\
\hline
\multicolumn{2}{l}{\footnotesize FD represent the frame difference signal} \\
\end{tabular}}
\end{table}

\newpage
\bibliography{references}

@article{levy2017impact,
  title={Impact of hands-on care on infant sleep in the neonatal intensive care unit},
  author={Levy, Jennifer and Hassan, Fauziya and Plegue, Melissa A and Sokoloff, Max D and Kushwaha, Juhi S and Chervin, Ronald D and Barks, John DE and Shellhaas, Ren{\'e}e A},
  journal={Pediatric pulmonology},
  volume={52},
  number={1},
  pages={84--90},
  year={2017},
  publisher={Wiley Online Library}
}

@article{haskova2014apnea,
  title={Apnea in preterm newborns: determinants, pathophysiology, effects on cardiovascular parameters and treatment},
  author={Haskova, K and Javorka, K and Javorka, M and Matasova, K and Zibolen, M},
  journal={Acta Medica Martiniana},
  volume={13},
  number={3},
  pages={5},
  year={2014},
  publisher={De Gruyter Poland}
}

@article{joosten2017sleep,
  title={Sleep related breathing disorders and indications for polysomnography in preterm infants},
  author={Joosten, Koen and de Goederen, Robbin and Pijpers, Angelique and Allegaert, Karel},
  journal={Early Human Development},
  volume={113},
  pages={114--119},
  year={2017},
  publisher={Elsevier}
}

@article{coleman2022assessment,
  title={Assessment of neonatal respiratory rate variability},
  author={Coleman, Jesse and Ginsburg, Amy Sarah and Macharia, William M and Ochieng, Roseline and Chomba, Dorothy and Zhou, Guohai and Dunsmuir, Dustin and Karlen, Walter and Ansermino, J Mark},
  journal={Journal of Clinical Monitoring and Computing},
  volume={36},
  number={6},
  pages={1869--1879},
  year={2022},
  publisher={Springer}
}

@article{villarroel2020non,
  title={Non-contact vital-sign monitoring of patients undergoing haemodialysis treatment},
  author={Villarroel, Mauricio and Jorge, Jo{\~a}o and Meredith, David and Sutherland, Sheera and Pugh, Chris and Tarassenko, Lionel},
  journal={Scientific reports},
  volume={10},
  number={1},
  pages={18529},
  year={2020},
  publisher={Nature Publishing Group UK London}
}

@article{honda2023effect,
  title={Effect of averaging time and respiratory pause time on the measurement of acoustic respiration rate monitoring},
  author={Honda, Jun and Murakawa, Masahiro and Inoue, Satoki},
  journal={JA Clinical Reports},
  volume={9},
  number={1},
  pages={61},
  year={2023},
  publisher={Springer}
}

@article{pergolizzi2022limited,
  title={The limited management options for apnoea of prematurity},
  author={Pergolizzi Jr, Joseph V and Fort, Prem and Miller, Thomas L and LeQuang, Jo Ann and Raffa, Robert B},
  journal={Journal of Clinical Pharmacy and Therapeutics},
  volume={47},
  number={3},
  pages={396--401},
  year={2022},
  publisher={Wiley Online Library}
}

@article{o2025caffeine,
  title={Caffeine and preterm infants: multiorgan effects and therapeutic creep: scope to optimise dose and timing},
  author={O’Shea, Michael and Butler, Luke and Holohan, Sam and Healy, Kate and O’Farrell, Rebecca and Shamit, Amreena and Cusack, Ruth and Elhadi, Mai and Lynch, Sinead and Gilcrest, Megan and others},
  journal={Pediatric Research},
  pages={1--8},
  year={2025},
  publisher={Nature Publishing Group US New York}
}

@phdthesis{sharma2025chronological,
  title={Chronological evaluation of functional changes in neonatal skin: A temporal evaluation of skin maturation},
  author={Sharma, Anushma},
  year={2025},
  school={University of Southampton}
}

@article{michaelis2025vitalcsi,
  title={{VitalCSI}: Contactless Respiratory Rate Estimation Using Consumer-Grade Wi-Fi Channel State Information},
  author={Michaelis, Tom and Jorge, Jo{\~a}o and Bijlani, Nivedita and Villarroel, Mauricio},
  journal={Sensors (Basel, Switzerland)},
  volume={26},
  number={1},
  pages={225},
  year={2025}
}

@book{world2023born,
  title={Born too soon: decade of action on preterm birth},
  author={{World Health Organization (WHO)}},
  year={2023},
  publisher={Geneva: World Health Organization},
 note={ Licence: CC BY-NC-SA 3.0 IGO}
  }

@article{koteswari2022preterm,
  title={Preterm birth: causes and complications observed in tertiary care hospitals},
  author={Koteswari, Poluri and Lakshmi, Pilly Aishwarya and Yaseen, Mohammed and Sultana, Sameera and Tabassum, Amena and Soumya, Paspula and Kawkab, Aasimah},
  journal={Cellular, Molecular and Biomedical Reports},
  volume={2},
  number={4},
  pages={202--2012},
  year={2022},
  publisher={Global Sciences Publisher}
}

@article{platt2014outcomes,
  title={Outcomes in preterm infants},
  author={Platt, MJ},
  journal={Public health},
  volume={128},
  number={5},
  pages={399--403},
  year={2014},
  publisher={Elsevier}
}

@article{thompson2024apnea,
  title={Apnea of Prematurity and Oxidative Stress: Potential Implications},
  author={Thompson, Lauren and Werthammer, Joseph W and Gozal, David},
  journal={Antioxidants},
  volume={13},
  number={11},
  pages={1304},
  year={2024},
  publisher={MDPI}
}

@article{joshi2016pattern,
  title={Pattern discovery in critical alarms originating from neonates under intensive care},
  author={Joshi, Rohan and van Pul, Carola and Atallah, Louis and Feijs, Loe and Van Huffel, Sabine and Andriessen, Peter},
  journal={Physiological Measurement},
  volume={37},
  number={4},
  pages={564--579},
  year={2016},
  publisher={IOP Publishing}
}

@article{serame2026contact,
  title={Deep learning-based detection of apnoea-related respiratory pauses in pre-term infants},
  author={Serame, Dineo and Villarroel, Mauro and Tarassenko, Lionel},
  journal={medRxiv},
  year={2026},
  doi={\textbf{Paper in preparation/archive}}
}

@article{moxon2015inpatient,
  title={Inpatient care of small and sick newborns: a multi-country analysis of health system bottlenecks and potential solutions},
  author={Moxon, Sarah G and Lawn, Joy E and Dickson, Kim E and Simen-Kapeu, Aline and Gupta, Gagan and Deorari, Ashok and Singhal, Nalini and New, Karen and Kenner, Carole and Bhutani, Vinod and others},
  journal={BMC pregnancy and childbirth},
  volume={15},
  number={Suppl 2},
  pages={S7},
  year={2015},
  publisher={Springer}
}

@article{kamala2022availability,
  title={Availability and functionality of neonatal care units in healthcare facilities in Mtwara region, Tanzania: The quest for quality of in-patient care for small and sick newborns},
  author={Kamala, Serveus Ruyobya and Julius, Zamoyoni and Kosia, Efraim M and Manzi, Fatuma},
  journal={Plos one},
  volume={17},
  number={11},
  pages={e0269151},
  year={2022},
  publisher={Public Library of Science San Francisco, CA USA}
}

@article{ahani2024video,
  title={Video-based respiratory rate estimation for infants in the NICU},
  author={Ahani, Soodeh and Niknafs, Nikoo and Lavoie, Pascal M and Holsti, Liisa and Dumont, Guy A},
  journal={IEEE Journal of Translational Engineering in Health and Medicine},
  year={2024},
  publisher={IEEE}
}

@article{bazarevsky2020blazepose,
  title={Blazepose: On-device real-time body pose tracking},
  author={Bazarevsky, Valentin and Grishchenko, Ivan and Raveendran, Karthik and Zhu, Tyler and Zhang, Fan and Grundmann, Matthias},
  journal={arXiv preprint arXiv:2006.10204},
  year={2020}
}

@article{ponsiglione2025comparison,
  title={Comparison of Techniques for Respiratory Rate Extraction from Electrocardiogram and Photoplethysmogram},
  author={Ponsiglione, Alfonso Maria and Russo, Michela and Petrellese, Maria Giovanna and Letizia, Annalisa and Tufano, Vincenza and Ricciardi, Carlo and Tedesco, Annarita and Amato, Francesco and Romano, Maria},
  journal={Sensors},
  volume={25},
  number={16},
  pages={5136},
  year={2025},
  publisher={MDPI}
}

@article{pinilla2025diagnostic,
  title={Diagnostic Modalities in Sleep Disordered Breathing: Current and Emerging Technology and Its Potential to Transform Diagnostics},
  author={Pinilla, Luc{\'\i}a and Chai-Coetzer, Ching Li and Eckert, Danny J},
  journal={Respirology},
  volume={30},
  number={4},
  pages={286--302},
  year={2025},
  publisher={Wiley Online Library}
}

@article{akbarian2020distinguishing,
  title={Distinguishing obstructive versus central apneas in infrared video of sleep using deep learning: Validation study},
  author={Akbarian, Sina and Montazeri Ghahjaverestan, Nasim and Yadollahi, Azadeh and Taati, Babak},
  journal={Journal of Medical Internet Research},
  volume={22},
  number={5},
  pages={e17252},
  year={2020},
  publisher={JMIR Publications Toronto, Canada}
}

@inproceedings{carter2023deep,
  title={Deep learning-enabled sleep staging from vital signs and activity measured using a near-infrared video camera},
  author={Carter, Jonathan and Jorge, Jo{\~a}o and Venugopal, Bindia and Gibson, Oliver and Tarassenko, Lionel},
  booktitle={Proceedings of the IEEE/CVF Conference on Computer Vision and Pattern Recognition},
  pages={5940--5949},
  year={2023}
}

@article{selvaraju2022continuous,
  title={Continuous monitoring of vital signs using cameras: A systematic review},
  author={Selvaraju, Vinothini and Spicher, Nicolai and Wang, Ju and Ganapathy, Nagarajan and Warnecke, Joana M and Leonhardt, Steffen and Swaminathan, Ramakrishnan and Deserno, Thomas M},
  journal={Sensors},
  volume={22},
  number={11},
  pages={4097},
  year={2022},
  publisher={MDPI}
}

@article{maurya2021non,
  title={Non-contact breathing monitoring by integrating {RGB} and thermal imaging via {RGB}-thermal image registration},
  author={Maurya, Lalit and Mahapatra, Prasant and Chawla, Deepak},
  journal={Biocybernetics and Biomedical Engineering},
  volume={41},
  number={3},
  pages={1107--1122},
  year={2021},
  publisher={Elsevier}
}

@article{ishrak2023doppler,
  title={Doppler radar remote sensing of respiratory function},
  author={Ishrak, Mohammad Shadman and Cai, Fulin and Islam, Shekh Md Mahmudul and Bori{\'c}-Lubecke, Olga and Wu, Teresa and Lubecke, Victor M},
  journal={Frontiers in Physiology},
  volume={14},
  pages={1130478},
  year={2023},
  publisher={Frontiers Media SA}
}

@article{ostojic2020reducing,
  title={Reducing false alarm rates in neonatal intensive care: a new machine learning approach},
  author={Ostojic, Daniel and Guglielmini, Sabino and Moser, Virginie and Fauch{\`e}re, Jean-Claude and Bucher, Hans Ulrich and Bassler, Dirk and Wolf, Martin and Kleiser, Stefan and Scholkmann, Felix and Karen, Tanja},
  journal={Oxygen Transport to Tissue XLI},
  pages={285--290},
  year={2020},
  publisher={Springer}
}

@article{zou2023new,
  title={A new approach to streamline obstructive sleep apnea therapy access using peripheral arterial tone-based home sleep test devices},
  author={Zou, Ding and Vits, Steven and Egea, Carlos and Ehrsam-Tosi, Daniela and Lavergne, Florent and Azpiazu, Mikel and Fietze, Ingo},
  journal={Frontiers in Sleep},
  volume={2},
  pages={1256078},
  year={2023},
  publisher={Frontiers Media SA}
}

@article{nwaneri2024review,
  title={A review of infant apnea monitor design},
  author={Nwaneri, Solomon and Ezenwa, Beatrice and Osuntoki, Akinniyi and Ezeaka, Veronica and Ogunsola, Folasade},
  journal={Journal of Clinical Sciences},
  volume={21},
  number={2},
  pages={93--98},
  year={2024},
  publisher={Medknow}
}

@article{poets2003pulse,
  title={Pulse oximetry vs. transcutaneous monitoring in neonates: practical aspects},
  author={Poets, Christian F},
  journal={www. bloodgas. com, Neonatology. Copenhagen: Radiometer Medical A/S},
  year={2003}
}

@InProceedings{Liu_2022_CVPR,
    author    = {Liu, Zhuang and Mao, Hanzi and Wu, Chao-Yuan and Feichtenhofer, Christoph and Darrell, Trevor and Xie, Saining},
    title     = {A ConvNet for the 2020s},
    booktitle = {Proceedings of the IEEE/CVF Conference on Computer Vision and Pattern Recognition (CVPR)},
    month     = {June},
    year      = {2022},
    pages     = {11976-11986}
}

@article{zarei2020automatic,
  title={Automatic classification of apnea and normal subjects using new features extracted from HRV and ECG-derived respiration signals},
  author={Zarei, Asghar and Asl, Babak Mohammadzadeh},
  journal={Biomedical Signal Processing and Control},
  volume={59},
  pages={101927},
  year={2020},
  publisher={Elsevier}
}

@article{yang2016sleep,
  title={Sleep apnea detection via depth video and audio feature learning},
  author={Yang, Cheng and Cheung, Gene and Stankovic, Vladimir and Chan, Kevin and Ono, Nobutaka},
  journal={IEEE Transactions on Multimedia},
  volume={19},
  number={4},
  pages={822--835},
  year={2016},
  publisher={IEEE}
}

@article{vitazkova2024advances,
  title={Advances in respiratory monitoring: a comprehensive review of wearable and remote technologies},
  author={Vitazkova, Diana and Foltan, Erik and Kosnacova, Helena and Micjan, Michal and Donoval, Martin and Kuzma, Anton and Kopani, Martin and Vavrinsky, Erik},
  journal={Biosensors},
  volume={14},
  number={2},
  pages={90},
  year={2024},
  publisher={MDPI}
}

@article{li2007robust,
  title={Robust heart rate estimation from multiple asynchronous noisy sources using signal quality indices and a Kalman filter},
  author={Li, Qiao and Mark, Roger G and Clifford, Gari D},
  journal={Physiological measurement},
  volume={29},
  number={1},
  pages={15},
  year={2007},
  publisher={IOP Publishing}
}

@phdthesis{lorato2021video,
  title={Video respiration monitoring: Towards remote apnea detection in the clinic},
  author={Lorato, Ilde Rosa},
  year={2021},
    school={Eindhoven University of Technology}
}

@article{lee2011new,
  title={A new algorithm for detecting central apnea in neonates},
  author={Lee, Hoshik and Rusin, Craig G and Lake, Douglas E and Clark, Matthew T and Guin, Lauren and Smoot, Terri J and Paget-Brown, Alix O and Vergales, Brooke D and Kattwinkel, John and Moorman, J Randall and others},
  journal={Physiological measurement},
  volume={33},
  number={1},
  pages={1},
  year={2011},
  publisher={IOP Publishing}
}

@article{massaroni2019contact,
  title={Contact-based methods for measuring respiratory rate},
  author={Massaroni, Carlo and Nicol{\`o}, Andrea and Lo Presti, Daniela and Sacchetti, Massimo and Silvestri, Sergio and Schena, Emiliano},
  journal={Sensors},
  volume={19},
  number={4},
  pages={908},
  year={2019},
  publisher={MDPI}
}

@article{bawua2021review,
  title={A review of the literature on the accuracy, strengths, and limitations of visual, thoracic impedance, and electrocardiographic methods used to measure respiratory rate in hospitalized patients},
  author={Bawua, Linda K and Miaskowski, Christine and Hu, Xiao and Rodway, George W and Pelter, Michele M},
  journal={Annals of Noninvasive Electrocardiology},
  volume={26},
  number={5},
  pages={e12885},
  year={2021},
  publisher={Wiley Online Library}
}

@article{pan1985real,
  title={A real-time QRS detection algorithm},
  author={Pan, Jiapu and Tompkins, Willis J},
  journal={IEEE transactions on biomedical engineering},
  number={3},
  pages={230--236},
  year={1985},
  publisher={IEEE}
}

@article{pergolizzi2022epidemiology,
  title={The epidemiology of apnoea of prematurity},
  author={Pergolizzi Jr, Joseph V and Fort, Prem and Miller, Thomas L and LeQuang, Jo Ann and Raffa, Robert B},
  journal={Journal of clinical pharmacy and therapeutics},
  volume={47},
  number={5},
  pages={685--693},
  year={2022},
  publisher={Wiley Online Library}
}

@article{bonner2017there,
  title={There were more wires than him: the potential for wireless patient monitoring in neonatal intensive care},
  author={Bonner, Oliver and Beardsall, Kathryn and Crilly, Nathan and Lasenby, Joan},
  journal={BMJ innovations},
  volume={3},
  number={1},
  year={2017},
  publisher={BMJ Specialist Journals}
}

@inproceedings{fei2009thermal,
  title={Thermal vision for sleep apnea monitoring},
  author={Fei, Jin and Pavlidis, Ioannis and Murthy, Jayasimha},
  booktitle={Medical Image Computing and Computer-Assisted Intervention--MICCAI 2009: 12th International Conference, London, UK, September 20-24, 2009, Proceedings, Part II 12},
  pages={1084--1091},
  year={2009},
  organization={Springer}
}

@article{muhammad2019p016,
  title={P016 Using non-invasive thermal imaging for apnoea detection},
  author={Muhammad, Usman and Evans, Ruth and Saatchi, Reza and Kingshott, Ruth and Elphick, Heather},
  journal = {BMJ Open Respiratory Research},
  year    = {2019},
  volume  = {6},
  pages   = {A9--A10}
}

@article{geertsema2020automated,
  title={Automated non-contact detection of central apneas using video},
  author={Geertsema, Evelien E and Visser, Gerhard H and Sander, Josemir W and Kalitzin, Stiliyan N},
  journal={Biomedical Signal Processing and Control},
  volume={55},
  pages={101658},
  year={2020},
  publisher={Elsevier}
}

@article{jorge2022non,
  title={Non-contact physiological monitoring of post-operative patients in the intensive care unit},
  author={Jorge, Jo{\~a}o and Villarroel, Mauricio and Tomlinson, Hamish and Gibson, Oliver and Darbyshire, Julie L and Ede, Jody and Harford, Mirae and Young, John Duncan and Tarassenko, Lionel and Watkinson, Peter},
  journal={NPJ digital medicine},
  volume={5},
  number={1},
  pages={4},
  year={2022},
  publisher={Nature Publishing Group UK London}
}

@article{adjei2021new,
  title={New method to measure interbreath intervals in infants for the assessment of apnoea and respiration},
  author={Adjei, Tricia and Purdy, Ryan and Jorge, Jo{\~a}o and Adams, Eleri and Buckle, Miranda and Fry, Ria Evans and Green, Gabrielle and Patel, Chetan and Rogers, Richard and Slater, Rebeccah and others},
  journal={BMJ open respiratory research},
  volume={8},
  number={1},
  pages={e001042},
  year={2021},
  publisher={Archives of Disease in childhood}
}

@inproceedings{cattani2014wire,
  title={A wire-free, non-invasive, low-cost video processing-based approach to neonatal apnoea detection},
  author={Cattani, L and Alinovi, D and Ferrari, Giorgio and Raheli, R and Pavlidis, E and Spagnoli, C and Pisani, F},
  booktitle={2014 IEEE Workshop on Biometric Measurements and Systems for Security and Medical Applications (BIOMS) Proceedings},
  pages={67--73},
  year={2014},
  organization={IEEE}
}

@inproceedings{zong2003open,
  title={An open-source algorithm to detect onset of arterial blood pressure pulses},
  author={Zong, W and Heldt, T and Moody, GB and Mark, RG},
  booktitle={Computers in Cardiology, 2003},
  pages={259--262},
  year={2003},
  organization={IEEE}
}

@article{lloyd2015overcoming,
  title={Overcoming the practical challenges of electroencephalography for very preterm infants in the neonatal intensive care unit},
  author={Lloyd, RO and Goulding, RM and Filan, PM and Boylan, GB},
  journal={Acta paediatrica},
  volume={104},
  number={2},
  pages={152--157},
  year={2015},
  publisher={Wiley Online Library}
}

@article{baharestani2007overview,
  title={An overview of neonatal and pediatric wound care knowledge and considerations.},
  author={Baharestani, Mona Mylene},
  journal={Ostomy/wound management},
  volume={53},
  number={6},
  pages={34--6},
  year={2007}
}

@article{lu2006semi,
  title={A semi-automatic method for peak and valley detection in free-breathing respiratory waveforms},
  author={Lu, Wei and Nystrom, Michelle M and Parikh, Parag J and Fooshee, David R and Hubenschmidt, James P and Bradley, Jeffrey D and Low, Daniel A},
  journal={Medical physics},
  volume={33},
  number={10},
  pages={3634--3636},
  year={2006},
  publisher={Wiley Online Library}
}

@article{zhao2011apnea,
  title={Apnea of prematurity: from cause to treatment},
  author={Zhao, Jing and Gonzalez, Fernando and Mu, Dezhi},
  journal={European journal of pediatrics},
  volume={170},
  number={9},
  pages={1097--1105},
  year={2011},
  publisher={Springer}
}

@article{bertoni2019towards,
  title={Towards Patient-centered Diagnosis of Pediatric Obstructive Sleep Apnea—A Review of Biomedical Engineering Strategies},
  author={Bertoni, Dylan and Isaiah, Amal},
  journal={Expert review of medical devices},
  volume={16},
  number={7},
  pages={617--629},
  year={2019},
  publisher={Taylor \& Francis}
}

@article{pullano2017medical,
  title={Medical devices for pediatric apnea monitoring and therapy: past and new trends},
  author={Pullano, Salvatore Andrea and Mahbub, Ifana and Bianco, Maria Giovanna and Shamsir, Samira and Islam, Syed Kamrul and Gaylord, Mark S and Lorch, Vichien and Fiorillo, Antonino S},
  journal={IEEE reviews in biomedical engineering},
  volume={10},
  pages={199--212},
  year={2017},
  publisher={IEEE}
}

@article{mohr2015very,
  title={Very long apnea events in preterm infants},
  author={Mohr, Mary A and Vergales, Brooke D and Lee, Hoshik and Clark, Matthew T and Lake, Douglas E and Mennen, Anne C and Kattwinkel, John and Sinkin, Robert A and Moorman, J Randall and Fairchild, Karen D and others},
  journal={Journal of Applied Physiology},
  volume={118},
  number={5},
  pages={558--568},
  year={2015},
  publisher={American Physiological Society Bethesda, MD}
}

@ARTICLE{8681397,
  author={Jorge, João and Villarroel, Mauricio and Chaichulee, Sitthichok and Green, Gabrielle and McCormick, Kenny and Tarassenko, Lionel},
  journal={IEEE Journal of Biomedical and Health Informatics}, 
  title={Assessment of Signal Processing Methods for Measuring the Respiratory Rate in the Neonatal Intensive Care Unit}, 
  year={2019},
  volume={23},
  number={6},
  pages={2335-2346},
  doi={10.1109/JBHI.2019.2898273}}

@phdthesis{montoya2017non,
  title={Non-contact vital sign monitoring in the clinic},
  author={Villarroel,Mauricio},
  year={2017},
  school={University of Oxford}
}

@article{villarroel2019non,
  title={Non-contact physiological monitoring of preterm infants in the Neonatal Intensive Care Unit},
  author={Villarroel, Mauricio and Chaichulee, Sitthichok and Jorge, Jo{\~a}o and Davis, Sara and Green, Gabrielle and Arteta, Carlos and Zisserman, Andrew and McCormick, Kenny and Watkinson, Peter and Tarassenko, Lionel},
  journal={NPJ digital medicine},
  volume={2},
  number={1},
  pages={1--18},
  year={2019},
  publisher={Nature Publishing Group}
}

@article{charlton2017breathing,
  title={Breathing rate estimation from the electrocardiogram and photoplethysmogram: A review},
  author={Charlton, Peter H and Birrenkott, Drew A and Bonnici, Timothy and Pimentel, Marco AF and Johnson, Alistair EW and Alastruey, Jordi and Tarassenko, Lionel and Watkinson, Peter J and Beale, Richard and Clifton, David A},
  journal={IEEE reviews in biomedical engineering},
  volume={11},
  pages={2--20},
  year={2017},
  publisher={IEEE}
}

@phdthesis{chaichulee2018non,
  title={Non-contact vital sign monitoring of pre-term infants},
  author={Chaichulee, Sitthichok},
  year={2018},
  school={University of Oxford}
}

\end{document}